\documentclass[sigconf, nonacm]{acmart}
\setcopyright{none}

\usepackage{multirow}
\usepackage{nicefrac}
\usepackage{cleveref}
\DeclareMathOperator*{\argmin}{arg\,min}
\usepackage{subcaption}
\usepackage{graphicx}
\usepackage{makecell}

\newcommand{\blfootnote}[1]{%
  \begingroup
  \renewcommand\thefootnote{}%
  \footnote{#1}%
  \addtocounter{footnote}{-1}%
  \endgroup
}

\newcommand{\R}{\mathbb{R}}

\begin{document}

\title{Compressive Meta-Learning}

\author{Daniel Mas Montserrat}
\affiliation{%
  \institution{Stanford University}
  \city{Stanford}
  \country{California, USA}}
\email{dmasmont@stanford.edu}

\author{David Bonet}
\affiliation{%
  \institution{University of California, Santa Cruz}
  \city{Santa Cruz}
  \country{California, USA}}
\email{dbonet@ucsc.edu}

\author{Maria Perera}
\affiliation{%
  \institution{Polytechnic University of Catalonia}
  \city{Barcelona}
  \country{Catalonia, Spain}}
\email{maria.perera.baro@gmail.com}

\author{Xavier Giró-i-Nieto}
\affiliation{%
  \institution{Polytechnic University of Catalonia}
  \city{Barcelona}
  \country{Catalonia, Spain}}
\email{xavigiro.upc@gmail.com}

\author{Alexander G. Ioannidis}
\affiliation{%
  \institution{University of California, Santa Cruz}
  \city{Santa Cruz}
  \country{California, USA}}
\email{ioannidis@ucsc.edu}

\renewcommand{\shortauthors}{Mas Montserrat et al.}

\begin{abstract}
The rapid expansion in the size of new datasets has created a need for fast and efficient parameter-learning techniques. Compressive learning is a framework that enables efficient processing by using random, non-linear features to project large-scale databases onto compact, information-preserving representations whose dimensionality is independent of the number of samples and can be easily stored, transferred, and processed. These database-level summaries are then used to decode parameters of interest from the underlying data distribution without requiring access to the original samples, offering an efficient and privacy-friendly learning framework. However, both the encoding and decoding techniques are typically randomized and data-independent, failing to exploit the underlying structure of the data. In this work, we propose a framework that meta-learns both the encoding and decoding stages of compressive learning methods by using neural networks that provide faster and more accurate systems than the current state-of-the-art approaches. To demonstrate the potential of the presented Compressive Meta-Learning framework, we explore multiple applications---including neural network-based compressive PCA, compressive ridge regression, compressive k-means, and autoencoders.
\end{abstract}


\maketitle

\blfootnote{Preprint. Extended version of a paper accepted at KDD '25. Publisher version: \href{https://doi.org/10.1145/3711896.3736889}{doi:10.1145/3711896.3736889}. This version is prepared for arXiv and may differ from the published version.}

\section{Introduction}

Compressive learning (CL) \cite{gribonval2020sketching, gribonval2021compressive} allows for efficient learning on large-scale datasets by compressing a complete dataset into a single mean embedding, also referred to as the sketch, which acts as a vector of generalized
moments. Ideally, the mean embedding will contain all the necessary information in order to learn the desired parameters of the underlying data distribution. The decoding of the parameters from the sketch (i.e. the learning process) is typically framed as an inverse optimization problem.
 The non-linear projection that generates the mean embedding takes a set of $N$ samples of $d$ dimensions and compacts them into a unique vector of $m$ dimensions where $m \ll Nd$. Note that this differs from traditional dimensionality reduction techniques (e.g. autoencoders or principal component analysis), since, while common dimensionality reduction techniques project $N \times d$ samples into $N \times h$ vectors with $h < d$, the mean embedding provides a compact representation for the totality of a data set, mapping $N \times d$ samples to a unique $m$-dimensional vector. Such a framework is particularly useful to learn models from data sets without the need for accessing the original samples directly, but instead only using the $m$-dimensional embedding. For example, CL techniques have proved to be effective at capturing parameters for Gaussian Mixture Models (GMMs), k-means, and PCA \cite{gribonval2021compressive}, from massive datasets with orders of magnitude lower computational requirements.
Note that the term sketching is also used in other areas such as data streaming applications \cite{cormode2005improved, charikar2002finding} and numerical linear algebra \cite{woodruff2014sketching}, although related, here ``sketching'' has a different meaning and methods between compressive learning and data streaming or numerical linear algebra are not directly comparable (See Appendix for further details).

Traditional supervised learning (e.g. SGD-based techniques) commonly relies on performing multiple passes over the dataset and computing a loss for each sample. 
While accurate, this paradigm requires to have access to the raw data, can be computationally intensive, and privacy-preserving mechanisms can be difficult to incorporate. 
Compressive learning provides an alternative paradigm that is (1) memory efficient,  (2) computationally efficient, and (3) privacy-friendly. Compressive learning makes use of  \emph{linear sketching} where the computation of the sketch can be parallelized throughout massive databases and these mean embeddings can be easily updated and support the addition and removal of samples. 
Namely, two sketches can be merged by a simple addition (or averaging), and new samples can be added and removed through sketch addition and subtraction. Note that linear sketching relates to how the sketch can be updated and does not imply that only linear functions can be used to compute the sketches. 
Such a parallel and online nature allows to efficiently compress datasets into embeddings that can be easily updated without the need to re-access the raw samples and can be easily stored and shared. Because the dimensionality of the embedding is independent of the size of the dataset, the learning process that maps the dataset-level embedding into the predicted parameters can be done efficiently, even for large datasets. Finally, differential privacy can be easily incorporated within compressive learning methods by adding the appropriate noise into the dataset-level sketch. Once differential privacy has been added to the embedding, all further processing, including the prediction of parameters, will maintain the privacy guarantees.

Two important limitations are present in current compressive learning systems. First, if the non-linear mapping function that projects the dataset into the mean embedding is not properly designed, the parameters of interest might not be learned accurately \cite{gribonval2020sketching, schellekens2020compressive}. Second, the current learning techniques are designed for a specific set of learning problems (e.g. k-means) and do not adapt well to new tasks, making it necessary to design a new learning approach for each application.
In this work, we introduce a new framework, Compressive Meta-Learning, that addresses both limitations by replacing the sketching and learning (i.e. decoding) operations with neural networks that are meta-learned end-to-end. 
First, a neural network (Sketch Network) performs a non-linear sample-wise projection (replacing the traditional randomized projections), followed by an average pooling operation that collapses all sample-level embeddings into a unique utility-preserving dataset-level embedding. Then, a second neural network (Query Network) takes as input the generated sketch and outputs the desired parameters (e.g. k-means centroids). We refer to this complete system as Sketch-Query Network (SQNet). The proposed method has several advantages: (a) the end-to-end training ensures that the sketching function properly captures the necessary information within the mean embedding. (b) By jointly training the sketching and learning functions, the generated sketch is specifically tailored to the query network, allowing it to accurately predict the desired set of parameters, and (c) the system can be meta-learned to predict parameters from complex models (e.g. an autoencoder) by simply changing the loss function, a task currently not possible with traditional compressive learning.

\section{Related Work}

The term ``sketching'' is used in multiple areas, and while all share a common theme of general purpose dimensionality reduction, they have different characteristics depending on the field.

\paragraph{Compressive Learning} Sketching techniques are used in compressive learning to project a dataset into a single vector (sketch) which captures the necessary information to learn the parameters of a model. In other words, a sketching function $f$, maps the dataset into a sketch $f \colon \R^{N\times d} \rightarrow \R^{m}$, and then a decoding function, $g$, maps the sketch into the parameter space $g \colon \R^{m} \rightarrow \R^{q}$, where $q$ is the dimensionality of the parameters. It is common that $m > d$ and $m \ll Nd$. For example, in compressive k-means, the decoding function maps the sketch $z \in \R^{m}$ into the $k$ cluster centroids $\theta \in \R^{k \times d}$. Most CL-based applications make use of Random Fourier Features (RFFs) \cite{rahimi2007random} to project each sample into a higher-dimensional space, and a pooling average is performed to obtain a dataset-level descriptor. The mapping from the sketch and the parameters of interest is typically framed as an inverse optimization problem, such as CL-OMPR \cite{keriven2018sketching, keriven2017compressive} and CL-AMP \cite{byrne2019sketched}, where the predicted parameters are iteratively updated by minimizing the error between the sketch computed with the original data and an empirical sketch computed from the predicted parameters. 
Some examples of CL applications include compressive k-means \cite{keriven2017compressive, schellekens2018quantized}, Compressive Gaussian Mixture Models \cite{keriven2018sketching},
Compressive PCA \cite{gribonval2020sketching, gribonval2021compressive}, linear regression \cite{gribonval2021compressive, dass2021householder}. Compressive multi-class classification \cite{schellekens2018compressive}, and generative network training \cite{schellekens2020compressive,perera2022generative}. Differential privacy has been successfully applied within CL applications \cite{chatalic2022compressive}. Recent works have explored using Nystrom approximations \cite{chatalic2022mean} to generate the sketches as an alternative to random features.

\paragraph{Data streaming} Data sketching has been widely applied in streaming applications \cite{cormode2013summary, cormode2017data}, where many sketching methods have been developed to approximately capture the frequency or presence of items, quantiles, or distinct counts of high-dimensional datastreams. Some methods include Count-Min \cite{cormode2005improved}, Count-Sketch \cite{charikar2002finding}, Bloom Filters \cite{bloom1970space}, HyperLogLog \cite{flajolet2007hyperloglog}, AMS Sketch \cite{alon1999space}, and Tensor Sketch \cite{pagh2013compressed}, which rely on hashing and sketching via random and sparse projections to map very high-dimensional vectors into compact representations that allow decoding important count-related information. Namely these techniques implement a mapping from $N \times d$ dimensional data into a compact sketch of dimension $m$, with $f \colon \R^{N\times d} \rightarrow \R^m$, $m \ll d$, and large dimensionality ($d$) and sample size ($N$). Such techniques include a decoding function, typically based on inverse linear projections and heuristics, to map the $m$-dimensional sketch into some $q$-dimensional representation. Recent works have incorporated supervised learning \cite{hsu2019learning, aamand2019learned,kraska2018case}.

\paragraph{Numerical linear algebra (NLA)}  Sketching techniques are used in applications including linear regression, PCA, and matrix factorization, among others. Typically, a randomized projection is used to reduce the dimensionality of matrices by combining rows (or columns) in order to obtain faster computations, namely $f \colon \R^{N\times d} \rightarrow \R^{l\times d}$, where a matrix $A \in \R^{N\times d}$ is projected with the projection $S \in \R^{l\times N}$ to obtain a compact representation $B = SA$, $B \in \R^{l\times d}$, with $l \ll N$. In many cases, an approximation of $A$ can be recovered from the compact representation $B$. The projection $S$ will typically be selected such as $||B^TB|| \approx ||A^TA||$ or $||Bx|| \approx ||Ax||$ for a given $x$, with theoretical guarantees that ensure that the sketch is a good approximation with probabilistic bounds on the loss of accuracy. Some examples include the Fast Johnson-Lindenstrauss Transform (FJLT) \cite{ailon2009fast}, randomized singular value decomposition (SVD) \cite{drineas2016randnla}, or randomized range finder for low-rank matrix approximation \cite{halko2009finding}, among many others \cite{woodruff2014sketching}. 
Recent works on sketching-based NLA have explored learning the sketching projections \cite{indyk2019learning, indyk2021few, liu2020learning, liu2022tensor}. While NLA techniques are applicable to settings that can be framed as matrix decompositions or similar, our proposed framework is applicable to any learning task as long as a differentiable function can be defined (e.g. predicting weights of an autoencoder). 

\begin{figure*}[!ht]
  \centering
  \includegraphics[width=0.95\linewidth]{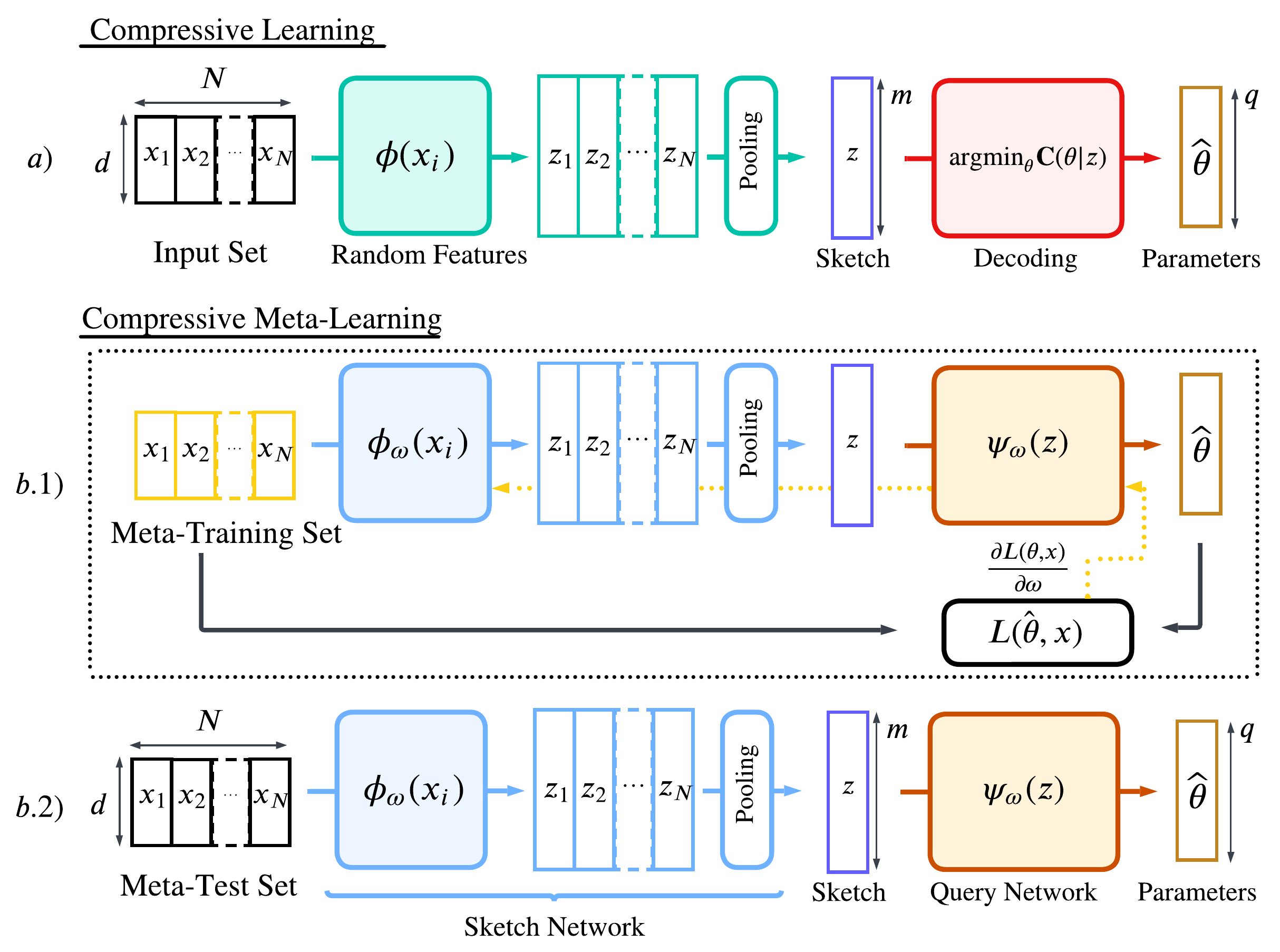}
  \caption{(a) Compressive Learning, (b) Compressive Meta-Learning with (b.1) Meta-training of Sketch-Query Network (SQNet), and (b.2) Meta-testing of SQNet.}
  \Description{Diagram of Compressive Meta-Learning}
  \label{fig:sqnet}
\end{figure*}

\paragraph{Deep Sets and Conditional Neural Processes}
Our work adapts a similar structure to the previously proposed Deep Sets \cite{zaheer2017deep}, which are neural networks that process each sample from a set and perform a permutation-invariant pooling operation for supervised and unsupervised applications. A permutation equivariant version of Deep Sets is also introduced in \cite{zaheer2017deep}. Deep Set Prediction Networks (DSPN) \cite{zhang2019deep} introduce an iterative neural network-based approach that allows to auto-encode sets. Transformer Set Prediction Network (TSPN) \cite{kosiorek2020conditional} extends DSPNs and produces a set-level summary that is fed to a transformer to make predictions for each element of the set.  Conditional Neural Processes (CNPs) \cite{garnelo2018conditional} follow a similar structure where a neural network is applied independently to each sample and all predicted embeddings are combined to obtain a dataset-level summary which is fed to a second-stage network that performs supervised regression or classification. 
We adopt a similar structure to such networks and frame it within the CL paradigm to learn parameters from sketches.

\paragraph{Meta-learning.} 
The notion of \emph{learning-to-learn} emerged early on with seminal contributions \cite{schmidhuber1987evolutionary, thrun1998learning}, which laid the groundwork for models that adapt their own learning strategies across tasks. Building on these foundations, modern meta-learning approaches focus on fast adaptation to novel tasks~\cite{hospedales2021meta}. For every new task, a model $P_\theta(y|x,\mathcal{S})$ is learned, where $y$ is the target, $x$ is the test input, and $\mathcal{S} = \{X,Y\}$ is the support set.
Metric-based learning methods such as Matching Networks \cite{vinyals2016matching} and Prototypical Networks \cite{snell2017prototypical} map a labelled support set $\mathcal{S}$ into an embedding space, where a distance is computed with the embedding of an unlabelled query sample to map it to its label. As in kernel-based methods, the model $P_\theta$ can be obtained through $P_\theta (y|x,\mathcal{S}) = \sum_{x_i,y_i\in \mathcal{S}} K_\theta (x,x_i)y_i$. 
Optimization-based methods such as Model-agnostic meta-learning (MAML) \cite{finn2017model} learn an initial set of model parameters and perform a further optimization through a function $f_{\theta(\mathcal{S})}$, where model weights $\theta$ are adjusted with one or more gradient updates given the support set of the task $\mathcal{S}$, i.e., $P_\theta (y|x,\mathcal{S}) = f_{\theta(\mathcal{S})} (x,\mathcal{S})$.
Recent works have explored the use of hypernetworks for meta-learning \cite{bonet2024hyperfast}.

\paragraph{Dataset Distillation.}
Compressive learning and compressive meta-learning are related to techniques for dataset distillation \cite{yu2023dataset}. The objective of dataset distillation (DD), also known as dataset condensation (DC), is to create a much smaller dataset consisting of synthetic samples that enable models trained on it to perform comparably to those trained on the full original dataset. Most dataset distillation techniques rely on generating pseudo-samples by using gradient-based optimization techniques that compute gradients from pre-trained neural networks \cite{wang2018dataset}. Techniques such as compressive k-means can be seen as special cases of dataset distillation.

\section{Compressive Meta-Learning}
\label{seq:neural sketching}

Supervised learning tries to find parameters $\theta$ of a model that minimizes a loss function $\mathcal{L}(\cdot)$ given a training dataset $\mathbf{x} = \{x_1, x_2, ..., x_N\}$:

\begin{equation}
\label{eq:supervised-loss}
\theta^* = \argmin_{\mathbf{\theta}} \mathcal{L}(\theta | \mathbf{x}) = \argmin_{\mathbf{\theta}} \sum_{i=1}^{N} \ell(\theta | x_i)
\end{equation}
where $ \mathcal{L}(\cdot)$ is a loss function (e.g. negative log-likelihood) evaluated at each training sample. The loss function and parameters will vary depending on the problem at hand. The parameters that minimize the loss function can be approximated by optimization techniques such as gradient descent or EM. Compressive learning (Figure~\ref{fig:sqnet}a) takes a different approach: instead of searching for parameters that minimize a given loss function with respect to the training samples, a surrogate loss function $\mathbf{C}(\theta | z)$ is used which depends on the sketch $z$ but not on the training dataset $\mathbf{x}$ directly. First, a sketch $z$ is computed by averaging per-sample non-linear projections:
\noindent\begin{minipage}{.53\linewidth}
\begin{equation}
z = \Phi(\mathbf{x}) = \frac{1}{N}\sum_{i=1}^{N}\phi(x_i)
\label{eq:cl-1}
\end{equation}
\end{minipage}%
\begin{minipage}{.46\linewidth}
\begin{equation}
\hat{\theta} = \argmin_{\mathbf{\theta}} \mathbf{C}(\theta | z)
\label{eq:cl-2}
\end{equation}
\end{minipage}

where a mapping function $\Phi(\cdot)$ takes as input a set of $N$ $d$-dimensional samples $\mathbf{x} = \{x_1, x_2, ..., x_N\}$, with $x_i \in \R^d$, performs a non-linear projection $z_i = \phi(x_i)$ (sketch projection) of each sample individually, obtaining a sample-level representation, and combines all of these into a global dataset-level embedding $z$, or sketch, with an average pooling (Eq. \ref{eq:cl-1}). Then, the estimated parameters  $\hat{\theta}$ are obtained through an optimization process (Eq. \ref{eq:cl-2}) that minimizes a surrogate cost function $\mathbf{C}(\cdot)$, which acts as a proxy to a supervised loss counterpart $\mathcal{L}(\cdot)$, but involves only the sketch $z$ and doesn't require access to the original dataset $\mathbf{x}$, and  
with $z, z_i \in \R^m$,  $\phi \colon \R^{d} \rightarrow \R^m$, $\Phi \colon \R^{N\times d} \rightarrow \R^m$, $\hat{\theta} \in \R^q$, where $q$ will vary depending on the application, and $\mathbf{C} \colon \R^{q} \times \R^{m} \rightarrow \R$. In most compressive learning approaches, $\phi$ consists of random feature projections, and the optimization procedure that obtains the parameters from the sketch (Eq. \ref{eq:cl-2}) is performed with techniques such as CL-OMPR and CL-AMP. This has two main disadvantages: First, if the random feature projection is not properly selected, the obtained sketch $z$ will not capture the necessary information in order to decode the parameters. Second, it can be challenging to find an appropriate cost function $\mathbf{C}(\cdot)$ with an adequate optimization procedure (Eq. \ref{eq:cl-2}) that approximates a given supervised loss $\mathcal{L}(\cdot)$ and accurately maps the sketch to the parameters. 
If the sketch has a large enough dimensionality, the sketching method $\phi(x_i)$ is properly designed, and the optimization problem in Eq. \ref{eq:cl-2} is accurately solved, one can expect that $\hat{\theta}_{CL} \approx \theta^*$. In fact, providing bounds on the difference between supervised learning and compressive learning parameter estimates is possible \cite{gribonval2020sketching, gribonval2021compressive}.

We introduce Compressive Meta-Learning, a new framework where both the sketching and decoding functions are replaced by parameterized neural networks which are learned end-to-end. The proposed ``Sketch-Query Network'' (SQNet) includes an encoding network (Sketch Network $\Phi_{\omega}$, Eq. \ref{eq:sketch-net}) that generates the information-preserving dataset summaries, and a decoding network (Query Network $\psi_{\omega}$, Eq. \ref{eq:query-net}) that maps sketches to parameters of interest (Figure~\ref{fig:sqnet}b):

\noindent\begin{minipage}{.6\linewidth}
\begin{equation}
z = \Phi_{\omega}(\mathbf{x}) = \frac{1}{N}\sum_{i=1}^{N}\phi_{\omega}(x_i)
\label{eq:sketch-net}
\end{equation}
\end{minipage}%
\begin{minipage}{.39\linewidth}
\begin{equation}
\hat{\theta} = \psi_{\omega}(z)
\label{eq:query-net}
\end{equation}
\end{minipage}

with $\phi_{\omega} \colon \R^{d} \rightarrow \R^m$, $\Phi_{\omega} \colon \R^{N\times d} \rightarrow \R^m$, and $\psi_{\omega} \colon  \R^{m} \rightarrow \R^{q}$. This approach removes the need of selecting appropriate random features and surrogate losses $\mathbf{C}(\cdot)$ and allows one to simultaneously learn both the sketching function and decoding function in a supervised end-to-end fashion, even for applications where a surrogate loss $\mathbf{C}(\cdot)$ does not exist.
The meta-parameters $\omega$ of the Sketch-Query Network are learned through a meta-training process that tries to minimize a given optimization problem: $\hat{\omega} = \argmin_{\mathbf{\omega}} \mathcal{L}^{M}(\omega | \mathbf{x})$.
Once the parameters of $\Phi_{\omega}(\cdot)$ and $\psi_{\omega}(\cdot)$ have been found, the sketch-query network pair can be used to infer (i.e. learn) the parameters of interest $\hat{\theta} = \psi_{\omega}(z) = (\psi \circ \Phi)_{\omega}(\mathbf{x})$. 
In practice, $\mathcal{L}^{M}(\omega | \mathbf{x})$ can be obtained by substituting the predicted parameters by the Sketch-Query Network $\hat{\theta} = (\psi \circ \Phi)_{\omega}(\mathbf{x})$ into the supervised learning loss (Eq. \ref{eq:supervised-loss}), and use backpropagation to learn $\omega$:
\begin{equation}
\mathcal{L}^{M}(\omega | \mathbf{x}) = \mathcal{L}(\hat{\theta} | \mathbf{x}) =  \sum_{i=1}^{N} \ell((\psi \circ \Phi)_{\omega}(\mathbf{x}) | x_i)
\label{eq:meta-loss}
\end{equation}
Alternatively, the loss in Eq. \ref{eq:meta-loss} can optimize a surrogate problem from which parameters of interest can later be recovered (e.g. predicting a covariance matrix from which PCA and Ridge regression can be obtained).

\begin{table}[htpb]
  \caption{Summary of Supervised, Compressive, and Compressive Meta-Learning.}
  \Description{A comparison of compressive meta‐learning with traditional learning methods}
  \label{table:cl-loss-summary}
  \centering
  \small
  {\setlength{\tabcolsep}{4pt}       
   \renewcommand{\arraystretch}{0.25} 
   \resizebox{\columnwidth}{!}{%
   \begin{tabular}{@{}l@{\hskip 8pt}c@{\hskip 0pt}c@{\hskip 0pt}c@{}}
     \toprule
     Framework                   & Mean Embedding                                     & Parameter Learning                                                              & Meta-Learning                                                       \\
     \midrule
     \makecell[l]{Supervised\\Learning}      & $-$                                                & $\displaystyle \theta^*=\argmin_{\theta}\sum_{i=1}^N\mathcal{L}(\theta\mid x_i)$ & $-$                                                                 \\
     \addlinespace
     \makecell[l]{Compressive\\Learning}     & $z=\frac{1}{N}\sum_{i=1}^N\phi(x_i)$               & $\displaystyle \hat\theta_{CL}=\argmin_{\theta}\mathbf{C}(\theta\mid z)$         & $-$                                                                 \\
     \addlinespace
     \makecell[l]{Compressive\\Meta‐Learning} & $z=\frac{1}{N}\sum_{i=1}^N\phi_\omega(x_i)$        & $\displaystyle \hat\theta_{SQ}=\psi_\omega(z)$                                   & $\displaystyle \hat\omega=\argmin_{\omega}\mathcal{L}^M(\omega\mid\mathbf{x})$ \\
     \bottomrule
   \end{tabular}
  }
  }
\end{table}

A key aspect of CL techniques is their applicability across different data distributions without the need of performing any training of the sketching and decoding mechanisms. In this work, we explore training SQNets that can generalize to unseen datasets. Specifically, for each proposed application, we (meta-)train the Sketch and Query networks with a set of datasets to obtain the meta-parameters $\omega$ (Figure~\ref{fig:sqnet}b.1) and then perform the evaluation by predicting parameters $\hat{\theta}$ in a new unseen set of datasets (Figure~\ref{fig:sqnet}b.2). The Sketch-Query Network pair can be understood as a (meta-)learned learning algorithm $\hat{\theta} = A_\omega(\mathbf{x})$ that predicts parameters given a training dataset. Table~\ref{table:cl-loss-summary} provides a comparison between supervised, compressive, and compressive meta-learning. More details of the training and evaluation setup are provided in the Appendix.

 \paragraph{Efficient and Online Learning.} The sketch can be easily updated by adding or removing the projection of new samples, making sketching-based learning an excellent framework for online learning applications. The computational time to obtain the parameters from a sketch is independent of the dataset size and only depends on the complexity of the Query Network.

\paragraph{Private Sketching.} Because only access to the sketch is needed, and not to the original data samples, sketching-based learning is a good approach when data cannot be shared due to privacy restrictions. Previous works \cite{chatalic2022compressive} have successfully explored incorporating approximate differential privacy (DP) into the sketch generation process. Here, we explore the use of this technique within our proposed Sketch-Query Network.
The ($\epsilon$, $\delta$)-DP sketch $z_{\epsilon,\delta}$ can be computed as: $z_{\epsilon,\delta} = \nicefrac{ \sum_{i=1}^{N}\phi'(x_i) + \xi}{N + \zeta}$
where  $\phi'(x_i) = \phi(x_i) \min(1,\frac{S}{||\phi(x_i)||})$ is a norm clipped version of the sketch projection,  $S = \max_{x}(||\phi(x)||)$ is the maximum L2 norm of the sketch projection across the meta-training samples, $\xi \in \R^m$, $\xi \sim \mathcal{N}(\mathbf{0},\,\sigma_1^{2}\mathbf{I}_m)$ is a Gaussian  noise, and $\zeta \sim \mathrm{Laplace}(\sigma_2)$ is a Laplacian noise, and $\sigma_1$ and $\sigma_2$ are selected as in \cite{chatalic2022compressive, balle2018improving}. 

DP ensures that any post-processing applied to $z_{\epsilon,\delta}$ will preserve its privacy guarantees. In practice, we perform a clamping of the sketch to remove potential out-of-range values. Then, we can use the query network to learn differentially private parameters: $\hat{\theta}_{\epsilon,\delta} = \psi(z^{c}_{\epsilon,\delta})$. In the experimental results, we show that by applying differential privacy during the meta-training process we learn Sketch and Query Networks that provide robust private estimates.
Both the dimensionality of the sketch and the number of samples used to compute it will have an important impact on how much information is preserved after adding differential privacy. Furthermore, one can easily show that:

\begin{equation}
\lim_{N\to\infty} z_{\epsilon,\delta} = z
\end{equation}

Therefore, as the number of samples used to generate the sketch increases, the distortion generated by the differential privacy becomes smaller. More details are provided in the Appendix.

\paragraph{Generalization Bounds.} The generalization properties of the parameters predicted by SQNet $\hat{\theta} = (\psi \circ \Phi)_{\omega}(\mathbf{x})$ are characterized by the complexity of the Sketch and Query networks. Specifically, the difference between the empirical $\mathcal{L}$ and expected error $\mathcal{L}^T$ of the predicted parameters can be bounded by the maximum norm of the sketch $\beta_\phi$ and loss function $M_\ell$,  the Lipschitz constants of the Query Network $\rho_\psi$ and of the loss $\rho_\ell$. Given a bounded sketching function one can show that the differences of sketches generated while replacing its $i$th sample is:

\begin{align}
||\Phi(\mathcal{S}) - \Phi(\mathcal{S}^{i})|| &= || \frac{1}{N}\sum_{x_j \in \mathcal{S}}\phi(x_j)  -  \frac{1}{N}\sum_{y_j \in \mathcal{S}^{i}}\phi(y_j)|| \\
&= ||\frac{1}{N} \phi(x_i)  -  \frac{1}{N}\phi(y_i)|| \\
&\leq  \frac{2\beta_\phi}{N}
\end{align}

By considering an $\rho_\psi$-Lipschitz Query Network, and a $\rho_\ell$-Lipschitz loss function, it follows that $\forall \mathcal{S}, \forall i$:

\begin{align}
\;\; ||\ell((\psi \circ \Phi)(\mathcal{S}),\cdot) - \ell((\psi \circ \Phi)(\mathcal{S}^{\backslash i}),\cdot)||_{\infty} & \leq \frac{\beta_\phi \rho_\psi \rho_\ell }{N}
\end{align}

Then, given uniform stability bounds \cite{bousquet2002stability} one can show that with probability $1-\delta$:

\begin{equation}
\mathcal{L}^T(\hat{\theta}) < \mathcal{L}(\hat{\theta}) + \frac{2\beta_\phi \rho_\psi \rho_\ell}{N} + (4\beta_\phi \rho_\psi \rho_\ell + M_\ell)\sqrt{\frac{\ln{\nicefrac{1}{\delta}}}{2N}}
\end{equation}

Intuitively, these results show that the generalization capabilities of the parameters learned with compressive meta-learning are proportional to the complexity of the decoding function (Query Network). If the Query Network has a small norm (leading to a small Lipschitz constant), the predicted parameters are guaranteed to generalize. The full proof is provided in the Appendix.

\begin{figure}[htpb]
  \centering
  \includegraphics[width=1\linewidth]{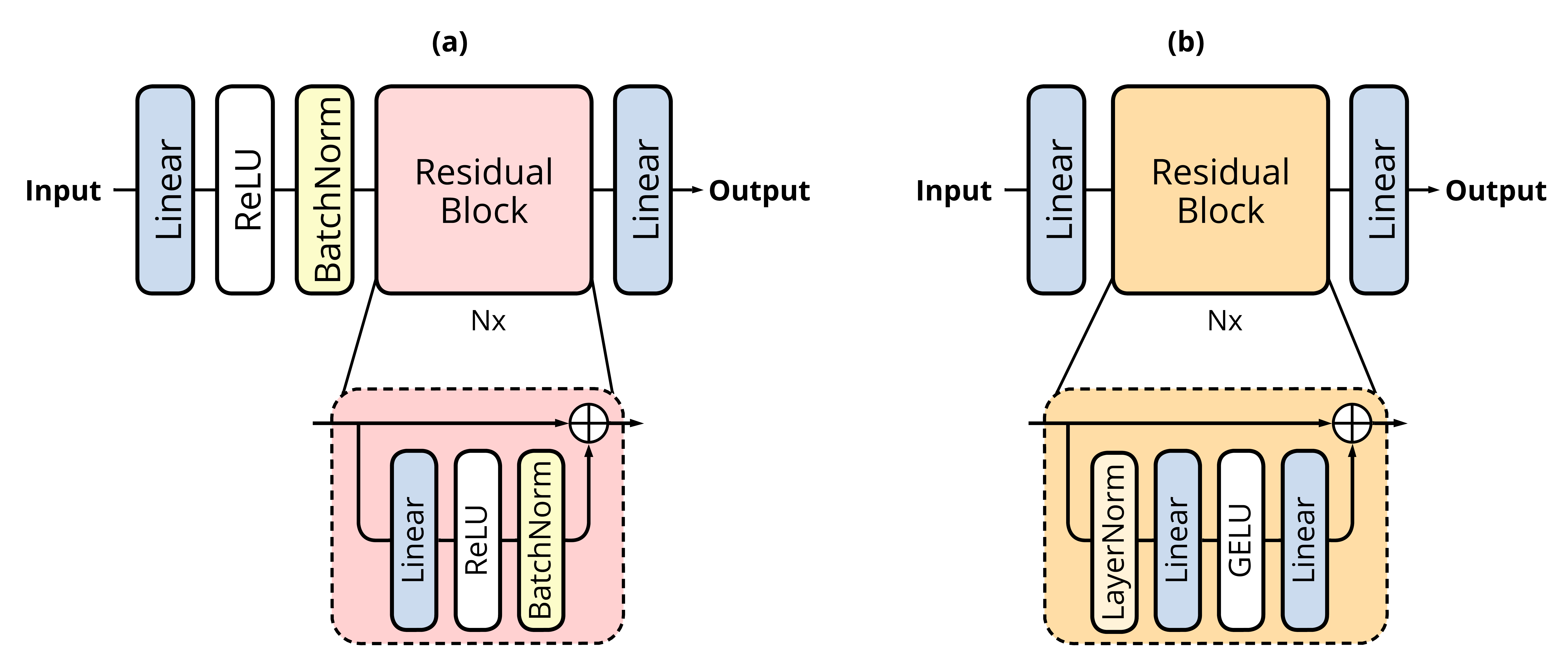}
  \caption{Different architectures used within SQNet.}
  \Description{Different architecture options with SQNet}
  \label{fig:net_arch}
\end{figure}

\paragraph{Neural Network Architectures.}
We make use of different residual networks as building blocks for Sketch Networks and Query Networks. Specifically we use a (a) residual batch-norm ReLU fully-connected network (``ResNet style'') and a (b) residual layer-norm GELU fully-connected network (``Transformer style''). Figure~\ref{fig:net_arch} provides a diagram for both types of architectures. When applied to the Sketch Network, a pooling layer is included at the end, and when applied to the Query Network, a sigmoid layer is applied if the application requires it. The only hyperparameters that we explore within the architecture is the number of residual blocks and the dimension of the hidden layers.

\section{Applications}
\label{sec:Applications}

\begin{table*}[t]
  \caption{Different applications with Supervised, Compressive, and Compressive Meta-Learning.}
  \Description{Comparison of applications with compressive meta-learning}
  \label{table:applications-loss-summary}
  \small
  \centering
  \begin{tabular}{lllll}
    \toprule
    Application     & Parameters  & \multirow{2}{11em}{Supervised Learning $\mathcal{L}(\theta | x_i)$} &  \multirow{2}{11em}{Compressive Learning $\mathbf{C}(\theta | z)$} &  \multirow{2}{11em}{Compressive Meta-Learning $\mathcal{L}^{M}(\omega | \mathbf{x})$}\\
     &   &  &  &  \\
    \midrule
    \multirow{2}{5em}{PCA}        & \multirow{2}{9em}{Orthonormal Basis \\ $\theta = \{\theta_1, ..., \theta_r \}$}  &  $|| x_j - \theta\theta^Tx_j ||^2 $  &  $||A\text{vec}(R) - z ||^2$  &  $||\text{vec}_{\text{LT}}(R) - \psi( \Phi(\mathbf{x}))||_1$   \\
    &   &  &  &  \\
    \midrule
    \multirow{2}{5em}{Ridge Regression}  & \multirow{2}{9em}{Linear weights \\ $\theta \in \R^{m \times n}$} &  $|| x^{(y)}_j - \theta x^{(x)}_j ||^2 + \lambda ||\theta ||^2$  &  $||A\text{vec}(R) - z ||^2$  &  $||\text{vec}_{\text{LT}}(R) - \psi( \Phi(\mathbf{x}))||_1$   \\

    &   &  &  &  \\
   \midrule
    \multirow{2}{5em}{$k$-means}  & \multirow{2}{9em}{$k$ centroids $\theta_i \in \R^d$ \\ $\theta = \{\theta_1, ..., \theta_k \}$} &  $\min_{k} || x_j - \theta_k ||^2$  &  $|| \Phi(\mathbf{\theta}) - z ||^2$  &  \multirow{2}{9em}{(a) $\min_{k} || x_j - \theta_k ||^2$ \\ (b) $||x_i - \hat{\theta}_{\pi(i)}||^2$ } \\

    &   &  &  &  \\
    \\[-0.5em]
    \midrule
    \multirow{2}{5em}{Autoencoder}  & \multirow{2}{9em}{Neural network weights $\theta$} &  $|| x_j - (g \circ f)_\theta(x_j) ||^2 $  &  $ - $  &  $|| x_j - (g \circ f)_{\psi( \Phi(\mathbf{x}))}(x_j) ||^2$   \\

    &   &  &  &  \\
    \bottomrule
  \end{tabular}
\end{table*}

We explore PCA, k-means, ridge regression, and autoencoder learning. Table~\ref{table:applications-loss-summary} provides an overview of the different applications and their respective loss functions. Both PCA and ridge regression learn a linear projection that can be re-framed as learning the data covariance from a sketch. k-means, which finds $k$ prototypical elements, is re-framed as reducing the L2 distance between the sketch generated using the samples of the dataset, and a sketch generated using the centroids $\Phi(\theta)$. Autoencoders, which are learned by minimizing a reconstruction loss in the supervised learning training, can be directly learned by predicting the weights from a sketch. Note that tasks such as autoencoder weight prediction do not have a clear compressive learning framing, further showing the benefits of the proposed method: by meta-training the sketch-query network pair, a mapping from a sketch to parameters can be learned even if no compressive learning criterion is available.

\subsection{Principal Component Analysis and Regression}
\label{ssec:pca}

Principal Component Analysis (PCA) tries to find a linear projection $\theta$ that minimizes the following mean squared reconstruction error,

\begin{equation}
\mathbf{\theta} = \argmin_{\mathbf{\theta}} \sum_{j=1}^{N} || x_j - \theta\theta^Tx_j ||^2
\label{eq:pca}
\end{equation}

where $\theta$ is an orthonormal projection.
It is well known that the principal components projections can be found by a simple eigendecomposition of the empirical covariance matrix of the data $R = \theta D \theta^T$, where $R = \frac{1}{N}\sum_{i=1}^{N}x_i x_i^T$ and $D$ is a diagonal matrix with eigenvalues of $R$. 
Ridge linear regression tries to find a regularized linear mapping such that:

\begin{equation}
\mathbf{\theta} = \argmin_{\mathbf{\theta}} \sum_{j=1}^{N} || x^{(y)}_j - \theta x^{(x)}_j ||^2 + \lambda ||\theta ||_2^2
\label{eq:regression}
\end{equation}

where $x_j = [x^{(y)}_j, x^{(x)}_j]$ are the regression labels $x^{(y)}_j$ and input features $x^{(x)}_j$ of the $j$th sample concatenated,   $R = \begin{pmatrix}
R_{11} & R_{12}\\
R_{21} & R_{22}
\end{pmatrix}$,
and $\theta = R_{12}(R_{22} + \lambda I)^{-1}$.
Therefore $R$ is sufficient statistic to obtain the PCA projection and Ridge regression parameters.

\paragraph{Compressive PCA and Linear Regression.} In many scenarios, $R$ can have very high dimensionality. Compressive PCA (CPCA) and Compressive Ridge Regression (CRR) \cite{gribonval2020sketching, gribonval2021compressive} try to provide a more efficient alternative by using the following sketch (Eq. \ref{eq:comp-pca-sketch}) and decoding functions  (Eq. \ref{eq:comp-pca-loss}):

\begin{equation}
z = \frac{1}{N}\sum_{i=1}^{N}A\text{vec}(x_i x_i^T) =A\text{vec}(R)
\label{eq:comp-pca-sketch}
\end{equation}
\begin{equation}
\hat{R} = \argmin_{R} ||A\text{vec}(R) - z ||^2
\label{eq:comp-pca-loss}
\end{equation}

where $\text{vec}(\cdot)$ flattens the $d \times d$ matrix into a $d^2$ vector, and $A$ is a random matrix with dimensions $m \times d^2$, $z$ is the empirical sketch. In practice, the size of the sketch is smaller than the size of the covariance matrix $m \ll d^2$. $\hat{R}$ can be found by minimizing Eq. \ref{eq:comp-pca-loss} through any desired optimization procedure, or by computing the pseudo-inverse of the randomized projection: $\text{vec}(\hat{R}) = A^+z$. Because $R$ is a symmetric matrix, the sketching process (Eq. \ref{eq:comp-pca-sketch}) and optimization objectives (Eq. \ref{eq:comp-pca-loss}) can be framed by using only the vectorized lower (or upper) triangular elements of $R$, i.e. replacing $\text{vec}(\hat{R})$ by $\text{vec}_{\text{LT}}(\hat{R}) \in  \R^{\frac{d(d+1)}{2}}$.

\begin{figure*}[htpb]
  \centering
  \begin{minipage}{0.33\linewidth}
    \includegraphics[width=\linewidth]{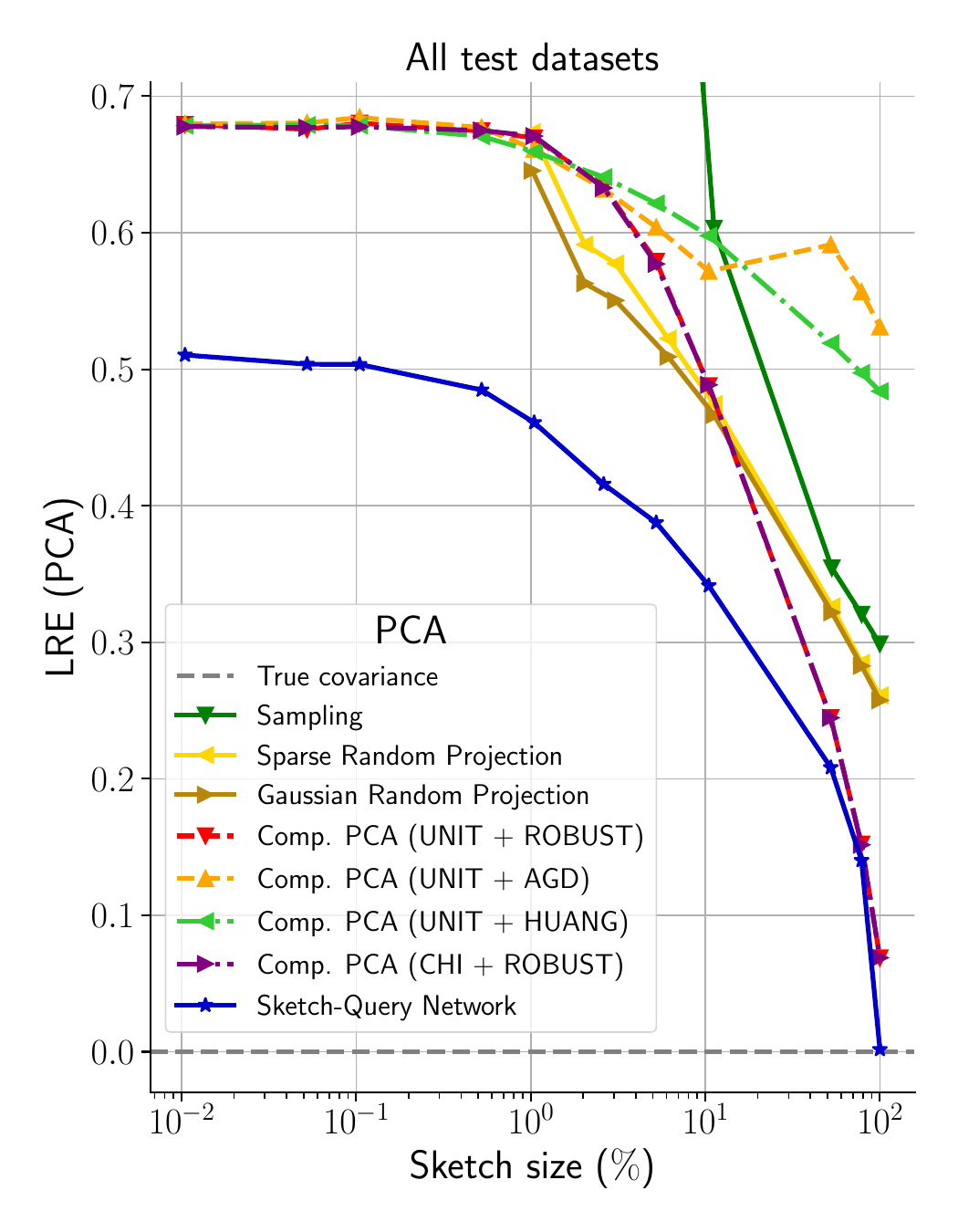}
  \end{minipage}
  \begin{minipage}{0.33\linewidth}
    \includegraphics[width=\linewidth]{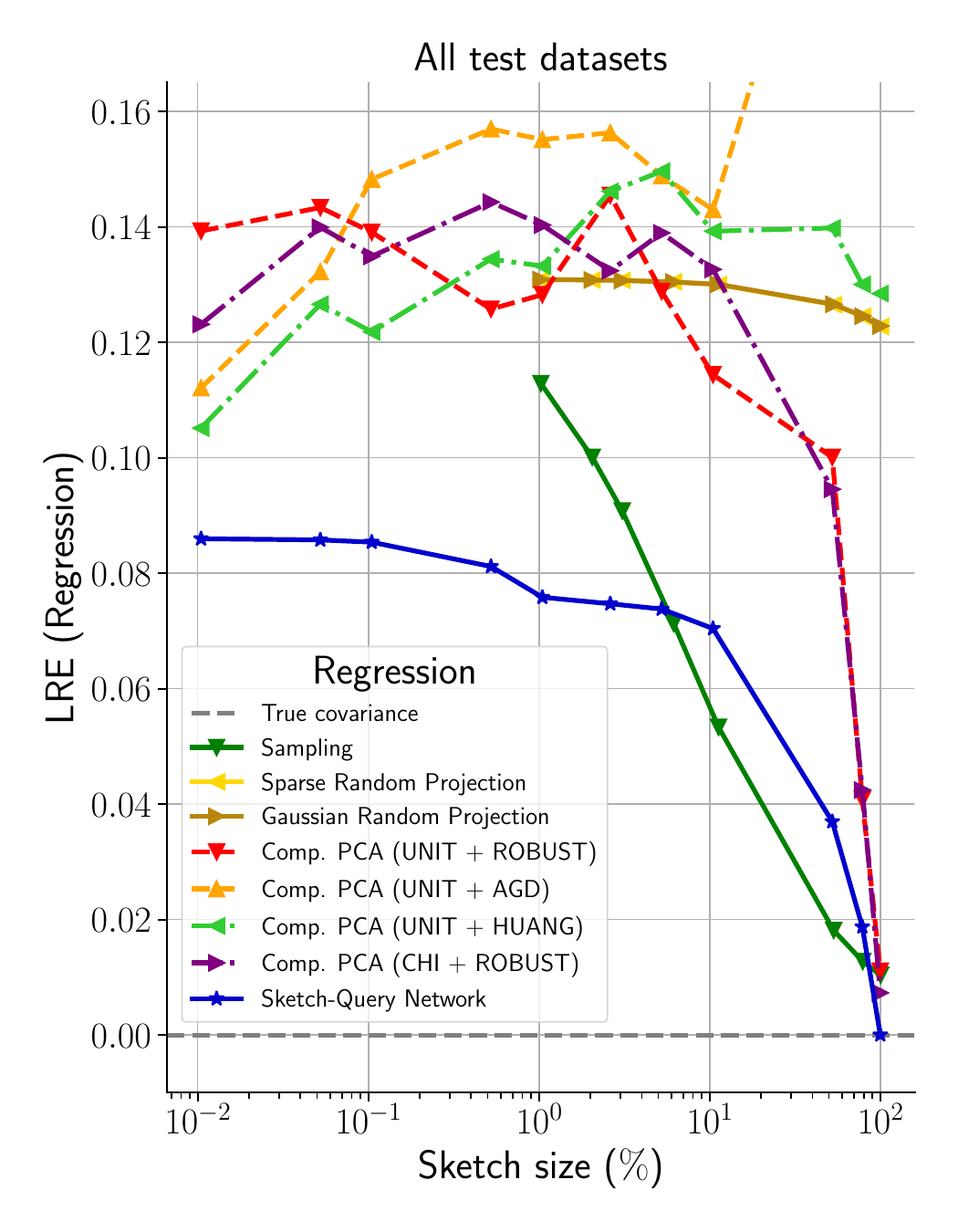}
  \end{minipage}
  \begin{minipage}{0.33\linewidth}
    \includegraphics[width=\linewidth]{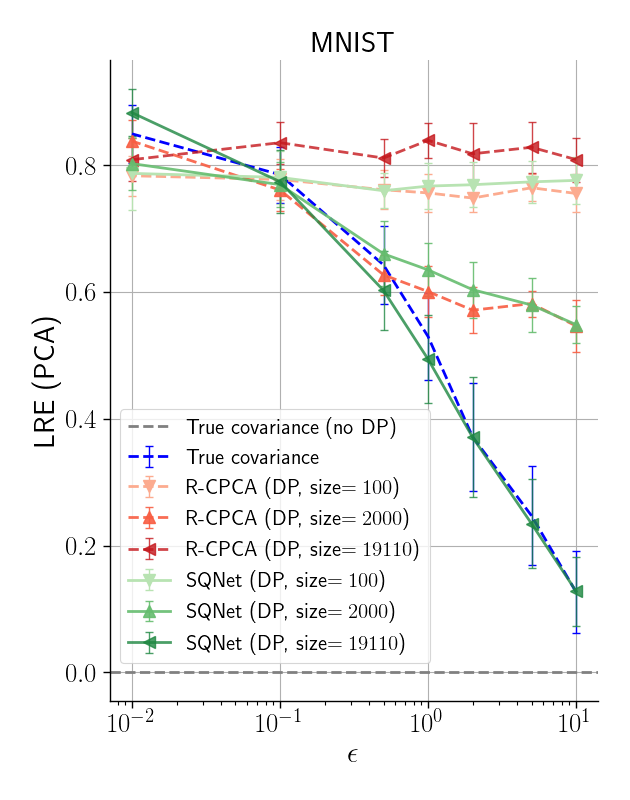}
  \end{minipage}
  \caption{(left) Logarithm relative scale error with respect to the true covariance error (LRE) of PCA reconstruction error, and (center) LRE of Regression for all datasets as a function of the sketch size (\% of the dimensionality of $R_{\text{LT}} \in \R^{d(d+1)/2}$). (right) LRE of PCA for the MNIST dataset incorporating differential privacy as a function of $\epsilon$.
  }
  \Description{Results on PCA}
\label{fig:pca_regression_results}

\end{figure*}

\paragraph{Neural-Based CPCA and CRR} We frame Compressive PCA and Compressive Ridge Regression as a Sketch-Query Network where both the sketch and the reconstructed covariance matrix are predicted with parametric models: $\text{vec}_{\text{LT}}(\hat{R}) = \psi( \Phi(\mathbf{x}))$.
The Sketch-Query Network pair is trained by minimizing the L1 error between the predicted and empirical covariance matrix:

\begin{equation}
\mathcal{L}(\mathbf{x}, \psi, \Phi) = ||\text{vec}_{\text{LT}}(R) - \psi( \Phi(\mathbf{x}))||_1
\label{eq:comp-pca-neural-loss}
\end{equation}

We train a sketch network consisting of a learned linear projection applied to the vectorized outer product of the input vector $\phi(x_i) = W_{\phi}\text{vec}(x_i x_i^T) + b_{\phi}$ followed by a query network consisting of a linear projection followed by a tanh activation $\psi(z) = \sigma(W_{\psi}z + b_{\psi})$. We train the network on a large range of datasets and evaluate it on new unseen datasets, showing that a learned Sketch-Query Network (Eq. \ref{eq:comp-pca-neural-loss}) can be used to learn PCA and regression parameters more accurately than traditional randomized compressive techniques (Eq. \ref{eq:comp-pca-loss}).

\paragraph{Experimental Results} We make use of the OpenML-CC18 suite \cite{bischl2021openml} composed of multiple datasets including tabular data and image datasets. We additionally use several MNIST-like image datasets including EMNIST Digits and Letters \cite{cohen2017emnist}, KMNIST \cite{clanuwat2018deep}, QuickDraw10 \cite{ha2018a}, and AfroMNIST \cite{wu2020afromnist}; DNA sequence datasets from Humans (HapMap3) \cite{international2010integrating} and Dogs (Canids) \cite{bartusiak2022predicting} are also included.
We randomly select 196 features from each dataset, and apply zero-padding to datasets with dimensionality smaller than 196. We perform a 50-50 split of datasets for meta-training and evaluation (detailed in the Appendix) and train the Sketch-Query Network with the meta-training split using Adam and a learning rate of $3 \times 10^{-5}$ with a learning rate scheduler, and compare it to traditional Compressive PCA \cite{gribonval2020sketching, gribonval2021compressive} approaches. 

To evaluate each method, we obtain the principal component projections, and the regression coefficients from the estimated covariance matrices for each dataset. Then, we compute the PCA reconstruction error (Eq. \ref{eq:pca}), and the regression MSE error, for all samples within the dataset. We repeat the process for the principal component dimensions (i.e. dimensionality of the projection) ranging from $1$ to $196$ and compute the average reconstruction error. We include baselines from the \textit{CompressiveLearning.jl} library \cite{chatalic2022mean}, namely Compressive PCA with random projections following the Chi distribution (CHI), projections whose columns are on the unit sphere (UNIT), and decoding methods including Robust Factorized Rank Minimization (ROBUST) \cite{gribonval2020sketching, gribonval2021compressive}, gradient-like approaches for sparse recovery (AGD) \cite{liu2018fast}, and Exponential-Type Gradient Descent Algorithm (HUANG) \cite{huang2018solving}. Additionally, we include baselines from the numerical linear algebra literature, including sparse and Gaussian projection, which combine rows of the dataset to reduce its dimensionality, and sampling, where a random subset of rows is selected \cite{woodruff2014sketching} (see Appendix). Figure~\ref{fig:pca_regression_results} (left) shows the average PCA reconstruction error for different sketch sizes ranging from 0.01\% to 100\% of the dimensionality of the covariance matrix. We can observe that the reconstruction error is consistently lower in Sketch-Query Network, outperforming all competing methods, and that it matches the reconstruction error of the actual PCA when the sketch dimensionality matches the dimensionality of $\text{vec}_{\text{LT}}(R)$. As the sketch becomes smaller, the difference between errors becomes larger, showing the importance of learning a good sketching mechanism. Figure~\ref{fig:pca_regression_results} (center) shows the regression reconstruction error, showing a similar trend as in PCA. Figure~\ref{fig:pca_regression_results} (right) shows the PCA error on the MNIST dataset when at different levels of differential privacy, showing that sketching can provide an accurate way to estimate private parameters, even surpassing the naive non-sketching-based approach. All results are reported in a logarithmic relative scale (LRE), with respect to the original error using the true covariance matrix $R$ (see the Appendix).

\begin{figure*}[htpb]
  \centering
  \includegraphics[width=0.95\linewidth]{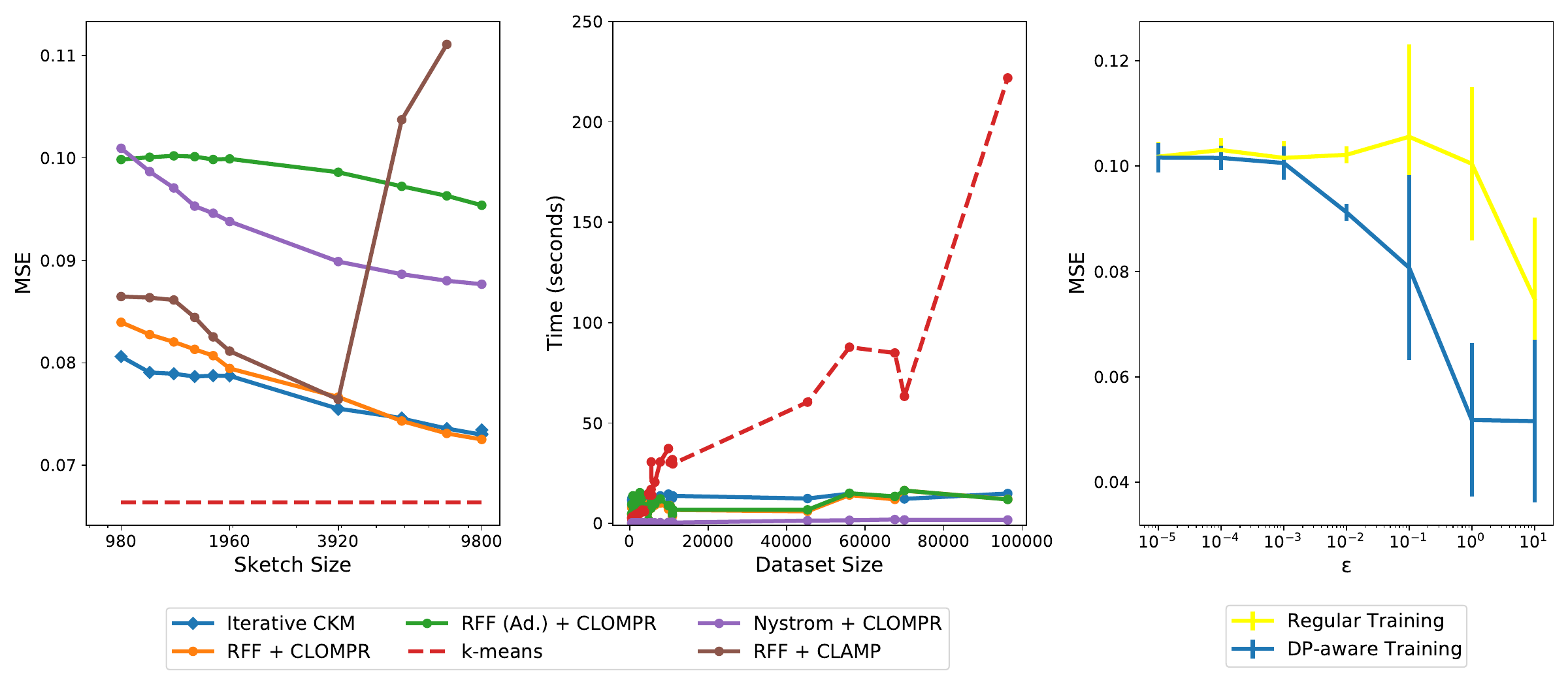}
  \caption{Benchmark of Compressive k-means methods with average MSE across datasets (left), computational time (center), and Iterative CKM results with DP-aware training for MNIST (right).}
  \Description{Results on k-means}
  \label{fig:kmeans-exp}
\end{figure*}

\subsection{k-means}
\label{ssec:kmeans}

k-means consists of finding $k$ centroids $\mathbf{\theta} = \{\theta_1, ..., \theta_K\}$, with $\theta_k \in \R^d$, such as the average mean square error between each training sample $x_j$ and its closest centroid $\theta_k$ is minimized:

\begin{equation}
\mathbf{\theta} = \argmin_{\mathbf{\theta}} \sum_{j=1}^{N} \min_{k} || x_j - \theta_k ||^2
\label{eq:kmeans}
\end{equation}

This is a widely used technique to perform unsupervised clustering and to learn cluster prototypes.

\paragraph{Compressive k-means.} As shown in \cite{keriven2017compressive, gribonval2020sketching, gribonval2021compressive} the centroids can be approximately found by minimizing the distance between the sketch of the dataset and the sketch generated using the centroids:

\begin{equation}
\hat{\mathbf{\theta}} = \argmin_{\mathbf{\theta}} || \Phi(\mathbf{\theta}) - \Phi(\mathbf{x}) ||^2 = \argmin_{\mathbf{\theta}} || \Phi(\mathbf{\theta}) - z ||^2
\label{eq:comp-kmeans}
\end{equation}

Previous works have successfully explored this Compressive k-means (CKM) approach by using Random Fourier Features (RFFs) to compute the sketch $z$ and using optimization techniques such as CL-OMPR to solve the objective in Eq. \ref{eq:comp-kmeans}. Such techniques are publicly available (e.g., at the \textit{CompressiveLearning.jl} library \cite{chatalic2022mean}).
The quality of the predicted k-means centroids $\hat{\theta}$ will depend on the projection function $\phi$ used to compute the sketch, and on the decoding algorithm used to map the sketch into predicted parameters. If either the projection function fails to properly capture the information of the underlying distribution of the data, or the decoding method fails to predict the parameters given the sketch, the quality of the prediction will be poor. Here we propose an iterative approach to learn a sketching $\Phi$ and decoding mechanism $\psi$.

\paragraph{Iterative CKM} 
By treating Eq. \ref{eq:comp-kmeans} as an iterative optimization process where we start with a set of random centroids $\mathbf{\hat{\theta}_{0}}$ and progressively update them to minimize the square distance between $\Phi(\mathbf{\hat{\theta}_{i}})$ and $z$, we can jointly optimize the sketching and query mechanisms. Namely, by performing the optimization using SGD and unfolding (unrolling) \cite{monga2021algorithm} the optimization procedure, the query network $\psi(z)$ can be formulated as:

\begin{equation}
\label{eq:kmeans-iter}
\mathbf{\hat{\theta}_{i+1}} = \mathbf{\hat{\theta}_{i}} - \alpha \nabla \mathcal{L}(\phi, \mathbf{\hat{\theta}_{i}}, z)
\end{equation}

where $\alpha$ is the learning rate, $\hat{\theta}_{i}$ are the estimated centroids in the current step, and $\mathcal{L}(\Phi(\mathbf{\hat{\theta}_{i}}), z) = || \Phi(\mathbf{\hat{\theta}_{i}}) - z ||^2_2$ is the mean square error loss between the sketch computed with the current centroids $\Phi(\mathbf{\hat{\theta}_{i}})$ and the empirical sketch $z$. Similar formulations can be pursued with other optimization algorithms, such as Adam \cite{kingma2014adam}. 
In fact, an unrolled optimization procedure defining the query function $\psi$ is equivalent to a recurrent neural network (RNN) defined by the gradient of the distance between sketches, and by training (learning) the sketching network ($\Phi$) the query network is simultaneously learned. 
Because training an unrolled optimization process, either through a differentiable optimizer, or implicit differentiation, can be unstable \cite{metz2021gradients}, we train this Sketch-Query Network with a derivative-free optimizer, the NGOpt optimizer \cite{bennet2021nevergrad}, as it provides us the flexibility to optimize simultaneously the weights of the network, the activation used, and hyperparameters of the inner optimization such as the inner learning rate (i.e. $\alpha$ in Eq. \ref{eq:kmeans-iter}), the inner optimizer used, which defines the iterative process in Eq. \ref{eq:kmeans-iter}, and the variance of the initial estimates $\hat{\theta}_0$. 
The proposed sketch network consists of a linear projection followed by a non-linearity $\phi(x) = \sigma(Wx)$. 

\paragraph{Experimental Results.} We used the same datasets and splits as in Compressive PCA, but we normalize all features to be bounded between 0 and 1. We compare the proposed Iterative CKM, the regular k-means, and traditional compressive k-means from \textit{CompressiveLearning.jl}, using Random Fourier Feature projections (RFF) \cite{keriven2018sketching, keriven2017compressive}, with and without adaptive radius (Ad.), Nystrom approximation \cite{chatalic2022mean}, and decoders such as CL-OMPR and CL-AMP. Figure~\ref{fig:kmeans-exp} (left) shows the mean square reconstruction error (MSE)  (Eq. \ref{eq:kmeans}) averaged across all testing datasets. For a fair comparison, the samples used in the Nystrom approximation are counted within the sketch size. Iterative CKM provides lower error than competing methods, especially with smaller sketch sizes, and matches compressive k-means with RFFs and CL-OMPR errors for larger sketch sizes. The error decreases as the sketch size increases, with methods based on CL-AMP providing unstable results. Figure~\ref{fig:kmeans-exp} (center) shows the computational time for each technique (excluding CL-AMP due to unstable results). As expected, compressive learning methods provide almost constant times regardless of the number of samples in the dataset, taking less than 20 seconds to process each dataset, while traditional k-means processing time grows super-linearly with the dataset size. Furthermore, we explore using DP within compressive k-means (Figure~\ref{fig:kmeans-exp} (right)). We use an Iterative CKM trained with (privacy-aware) and without  (regular) DP during training and evaluated on the MNIST dataset. Specifically, we apply an $(0.01,0.01)$-DP when performing privacy-aware training. We show that the sketching function learned using privacy-aware training provides lower reconstruction errors, specifically with values of $\epsilon$ close to $1$. With $\epsilon < 0.01$ both methods start performing poorly.

\subsection{Autoencoders}
\label{ssec:autoencoder}

An autoencoder (AE) combines an encoder that maps inputs $x$ into embeddings $u = f_{\theta}(x)$, and a decoder that tries to reconstruct the input $\hat{x} = g_{\theta}(u)$. The encoder-decoder pair $(g_\theta \circ f_\theta)(x_j)$ is parameterized by $\theta$ and learned by minimizing some reconstruction error such as:

\begin{equation}
\mathbf{\theta} = \argmin_{\mathbf{\theta}} \sum_{j=1}^{N} || x_j - (g_\theta \circ f_\theta)(x_j) ||^2
\label{eq:ae}
\end{equation}

Commonly, $\theta$ is estimated with an SGD-based method which can be slow, computationally intensive, and requires to have direct data access. Here we explore the application of SQNet to predict the parameters $\theta$ of AEs such that they can adapt to new, unseen datasets without the need for re-training the encoder-decoder pair, by replacing the slow training process of traditional AEs with the fast sketching and decoding to learn the parameters $\hat{\theta} = (\psi \circ \Phi)(\mathbf{x})$.

\paragraph{Sketch-Conditional Autoencoders.} The proposed AE has two sets of parameters: fixed parameters $v$, which are learned during the meta-training process and kept fixed afterwards, and dynamic parameters $\theta$, which are predicted from a sketch $z$ by the query network for every new dataset. The encoder, decoder, and sketch network consist of a residual MLP architecture, and the query network is a simple linear layer that transforms the sketch into the predicted dynamic weights. The dynamic weights $\theta$ of the encoder and decoder consist of the bias vectors of their respective first linear layer. Therefore, the output of the first linear layer of the encoder (and decoder), can be stated as:

\begin{equation}
h_1(x,z) = W_xx + \psi(z) = W_xx + W_zz + b
\end{equation}

where $W_x$ is the fixed (meta-learned) linear layer, and $\psi(z) = b(\mathbf{x}) = W_z z + b$ is the dynamic bias predicted from the sketch by the query network. By using skip connections, the information of the sketch can be propagated throughout all the layers. Note that this framework could be extended to predict more weights besides the dynamic biases.
The fixed weights of the AE $v$, the sketching $\Phi$, and query network $\psi$ are jointly learned during meta-training and kept fixed afterwards. The meta-learning is performed by predicting the sketch and dynamic bias with a batch of samples, and then computing the reconstruction error with a new batch from the same dataset. The error is backpropagated through the encoder, decoder, and Sketch-Query Networks. After training, when an unseen dataset $\mathbf{x'}$ is found, the dynamic biases are predicted by the Sketch-Query Network and introduced in the AE $b(\mathbf{x'}) = \psi(\Phi(\mathbf{x'}))$. Note that samples from the evaluation datasets are not used during the meta-training process. During testing, they are used both to generate the sketch and to evaluate the model's final performance.

We explore multiple variations of the autoencoder to properly assess the effect of conditioning by a learned sketch: a regular autoencoder without sketch conditioning (AE), trained with each of the evaluation datasets (i.e. dataset-specific AEs), an autoencoder conditioned with the sample mean (i.e. a sketch generated with the identity function) (+M), an autoencoder conditioned by the mean and a learned sketch (+MS), and an autoencoder conditioned by a per-class sketch, where a unique sketch is created for each class (+MSK). For a more detailed discussion, see Appendix.

\paragraph{Experimental Results.} We make use of multiple datasets: 2 datasets of images including MNIST (M) \cite{lecun1998mnist}, QuickDraw-10 (QD) \cite{ha2018a}, 2 datasets of genomic data, including human whole-genome (H) \cite{perera2022generative} and dogs (D) \cite{bartusiak2022predicting}, UCI datasets including KDD Cup 1998 (K) \cite{bay2000uci} and Adult (A) \cite{kohavi1996scaling}, and Kaggle's 
Bank Marketing (B) \cite{moro2014data} dataset. We process all datasets with binarization and one-hot encoding and randomly keep 1000 dimensions. 
The sketch-conditional AEs are meta-trained with a randomized binarized MNIST (RM) dataset, where the position of each pixel is shuffled and randomly negated at every batch. This heavily randomized data augmentation allows us to learn a network with generalization capabilities and forces the AE to extract useful information from the sketches. All AEs have a bottleneck dimension of 50.
As shown in Table~\ref{tab:autoencoder}, regular AEs are able to reconstruct with high accuracy in-distribution samples but completely fail to reconstruct out-of-distribution samples, with the exception of AEs trained with the image datasets (MNIST, QuickDraw) which can partly generalize to the other image datasets. On the other hand, the AE trained with randomized MNIST has higher average generalization capabilities, and it is surpassed by the mean-conditional,  sketch-conditional, and categorical-sketch-conditional networks. In almost all datasets, the per-class sketch-conditional AE (AE+MSK) provides the second-best reconstruction, in some cases with an accuracy comparable to the dataset-specific AE.

\begin{table}[!t]
  \centering
  \fontsize{9}{11}\selectfont
  \caption{Balanced accuracy of the AE reconstructions. TS: Train set; RM: Randomized MNIST; +M: mean-conditional; +S: sketch-conditional; +K: mean and sketch-conditional per class. \textbf{Bold} indicates best, \textcolor{blue}{Blue} second-best.}
  \begin{tabular}{l@{\hspace{4pt}}l@{\hspace{3pt}}|@{\hspace{6pt}}c@{\hspace{6pt}}c@{\hspace{6pt}}c@{\hspace{6pt}}c@{\hspace{6pt}}c@{\hspace{6pt}}c@{\hspace{6pt}}c@{\hspace{6pt}}|c}
    \toprule
    Model & TS   & M &  H  &  D   &  K  & QD  &  A & B  & Avg \\
    \midrule
    AE & M  & \textbf{99.7}   &  54.1   &  53.1 &  50.8    &  74.7 & 58.6 & 55.8 & 63.8    \\
    AE & H  & 54.8   &  \textbf{85.1}   &  54.5 & 56.4   &  54.7 & 58.4 & 57.6 & 60.2  \\
    AE & D  & 55.2   &  53.4   &  \textbf{80.4} &  52.5    &  55.2 & 56.7 & 58.1 & 58.8   \\
    AE & K  & 48.0    &  53.4  &  52.5 &  \textbf{88.4}   &  48.6 & 66.6 & 58.6 & 59.4    \\
    AE & QD  & \textcolor{blue}{95.6} & 53.7 & 53.1 & 51.4 & \textbf{84.5} & 55.3 & 52.4 & 63.7   \\
    AE & A  & 55.8  &  53.0   &  51.9 &  51.5  &  55.2 & \textbf{99.5} & 62.5 & 61.3  \\
    AE & B  & 51.2  & 50.7 & 50.8 & 50.5 & 50.6 & 62.0 & \textbf{99.2} & 59.3  \\
    \midrule
    AE &  RM  & 66.8  & 65.5 & 60.6 & 64.2 & 67.6 & 73.0 & 68.1 & 66.5   \\
    +M &  RM  & 78.6 & 80.5 & 70.2 & 72.6 & 59.5 & 76.8 & 72.3 & 72.9 \\
    +MS &  RM  & 79.2  & 81.2 & 70.1 & 73.7  & 61.9 & 80.2 & 81.5 & \textcolor{blue}{75.4}  \\
    +MSK &  RM  & 85.7  & \textcolor{blue}{82.4} & \textcolor{blue}{72.2} & \textcolor{blue}{77.0}  & \textcolor{blue}{68.3} & \textcolor{blue}{86.0} & \textcolor{blue}{87.5} & \textbf{79.9}  \\
    \bottomrule
  \end{tabular}
  \Description{More results on autoencoder reconstructions}
  \label{tab:autoencoder}
\end{table}

\section{Strengths and Limitations\\of Compressive Meta-Learning}

A key aspect of compressive learning is that it only requires a single pass through the dataset, making sketch computation linear with respect to the size of the dataset $N$, which can be fully parallelized. Furthermore, the computational complexity of predicting the parameters $\theta$ is independent of the dataset size. Such properties make compressive \mbox{(meta-)learning} a very good fit for privacy-preserving, online learning, or federated learning applications. 
However, the dimensionality of the sketch becomes an  important aspect in order to properly capture enough data for the successful decoding of the parameters. For example, to obtain an accurate reconstruction of the covariance matrix, a sketch of size $m$ with $d \leq m \leq d(d+1)/2$ is recommended. For k-means, a sketch size proportional to the dimensionality and $k$ is required with $m \propto kd$. This relationship between the input dimension $d$ and the sketch dimension $m$ can make it difficult to apply compressive learning-based techniques to high-dimensional data such as high-resolution images, text, 3D objects, or whole genome DNA sequences. Future works should tackle such challenges in order to provide a compressive \mbox{(meta-)learning} paradigm that can scale properly with the dimensionality of the data.

\section{Conclusions}
We have introduced meta-learning into compressive learning applications, demonstrating that neural networks can significantly improve accuracy and replace ad hoc randomized sketching and decoding mechanisms, while easily incorporating differential privacy.
Future work is required to apply compressive learning techniques to high-dimensional data. In many real-world applications, the sketch size needs to scale linearly, or even quadratically, with the input dimension, which poses a challenge. Although we show that our approach can be applied to data with dimensionality from hundreds to thousands, significantly larger than in previous works, new approaches are needed to handle many natural signals.

\bibliographystyle{ACM-Reference-Format}
\bibliography{main}

\appendix





\section{Compressive Meta-Learning Framework}

\subsection{Sketch Update and Pooling Operations}

Sketches can be easily updated: new samples can be added and removed from a sketch, and sketches can be combined or split. In order to merge two sketches $z_a$ computed with $N_a$ samples $\mathbf{x_a} = \{x_{a1}, x_{a2}, ..., x_{N_a}\}$ and $z_b$ computed with $N_b$ samples $\mathbf{x_b} = \{x_{b1}, x_{b2}, ..., x_{N_b}\}$, a new combined sketch can be computed as:

\begin{equation}
z' = \frac{N_a z_a + N_b z_b}{N_a + N_b}
\end{equation}

Note that in order to combine two sketches $z_a$ and $z_b$, the original samples $\mathbf{x_a}$ and $\mathbf{x_b}$ are not required, but only the sketches and their respective size $N_a$ and $N_b$. This makes the sketching framework well suited in online and federated learning applications, where new samples become available sequentially and can be continuously incorporated within the sketch, and in distributed and federated learning scenarios, where different parties compute sketches locally with their data, then share the sketches which are aggregated later on.

Similarly, sets of samples can be removed from sketches. To remove a set of $N_b$ samples $\mathbf{x_b} = \{x_{b1}, x_{b2}, ..., x_{N_b}\}$  from a sketch $z'$ computed with $N = N_a + N_b$ samples, a subtraction between sketches is required:

\begin{equation}
z_a = \frac{N z' - N_b z_b}{N}
\end{equation}

As in the sketch merging operation, the original samples are not required to subtract sketches, but only the computed sketches and their size. This type of operation can be useful in settings where the information of some samples needs to be forgotten. This can happen in medical applications, where patients decide to remove their data from databases, or in other services where users leave and request to have their data removed from databases and algorithms. While in traditional supervised learning, a new set of models and parameters might have to be trained without the removed data, sketching-based learning just requires one subtraction between sketches and a forward pass through the query network to obtain updated parameters.

We explore different pooling operations to combine the sample-level embeddings into a dataset-level sketch. Table~\ref{table:pooling} shows multiple layers and how new samples (or sketches) can be added and removed from the sketch. The \textit{mean} and \textit{summation} pooling are equivalent, up to a scaling factor, with the mean requiring the value of $N$, the total elements in the sketch. The \textit{max} operation does not allow for removal of samples; however, soft versions of the \textit{max} operation such as \textit{p-norm} (with a large enough p) and the \textit{log-sum-exp} can be used, which allow for addition and removal of samples. Note that due to rounding errors, removal of samples from \textit{p-norm} and \textit{log-sum-exp} layers can lead to noisy sketches. The \textit{min} pooling layer and its soft approximations can easily be applied by making use of the fact that $\min(a,b) = -\max(-a,-b)$. 
Furthermore, by re-defining the last layer of the Sketch Projection layer, and the first layer of the Query Network, all the pooling operations (except \textit{max} and \textit{min}) can be defined as a \textit{mean} pooling operation, assuming that $N$ is known (e.g. by considering $\phi'(x_i)=e^{\phi(x_i)}$ and  $\psi'(z)=\psi(\log(Nz))$, the \textit{log-sum-exp} can be replaced by a mean pooling layer). Therefore, the theoretical framework of CL is applicable regardless the adopted pooling layer. 

\begin{table}[ht]
  \caption{Multiple pooling operations.}
  \Description{Different pooling operations}
  \label{table:pooling}
  \small
  \centering
  \begin{tabular}{lll}
    \toprule
    Pooling Layer     & Merging  / Addition   & Removal \\
    \midrule
    Sum         & $z' = z_a + z_b $ &  $z' = z_a - z_b $    \\
    Mean        & $z' = \frac{N_a z_a + N_b z_b}{N_a + N_b} $ & $z' = \frac{N_a z_a - N_b z_b}{N_a - N_b} $   \\
    Max         & $z' = \max(z_a, z_b) $       & N/A  \\
    p-Norm      & $z' = \sqrt[p]{z_a^p + z_b^p} $      & $z' = \sqrt[p]{z_a^p - z_b^p} $  \\
    Log-Sum-Exp & $z' = \log(e^{z_a} + e^{z_b}) $       & $z' = \log(e^{z_a} - e^{z_b})$   \\
    \bottomrule
  \end{tabular}
\end{table}

\section{Extended Related Work}

Three of the main areas where sketching is found are data streaming, linear algebra, and compressive learning. Note that other works such as some computer vision tasks dealing with pictures of drawings might use the term sketching in a completely unrelated meaning and application.

\paragraph{Sketches for data streaming}

Data sketching has been widely applied in multimedia summarization, web and streaming applications, and data processing \cite{cormode2013summary, cormode2017data}. Such techniques must be computationally efficient and be able to handle constant streams of high-dimensional data.  Many sketching methods have been developed to approximately capture the frequency or membership of items, information about quantiles, or to count distinct elements. Some key methods include Count-Min \cite{cormode2005improved}, Count-Sketch \cite{charikar2002finding}, and Bloom Filters \cite{bloom1970space}. Count-Min performs multiple random projections using a sparse non-negative linear mapping. These random projections act as hashing functions to map the input elements into a lower dimensional vector (sketch) that keeps a compact count of the frequency of each item. A set of inverse linear projections are used to map the sketch into the estimated frequency per item. Each linear projection reconstructs a different set of frequencies and the final estimate can be obtained by selecting the minimum between them. Count-Sketch follows the same approach as Count-Min, but allows negative values in the random projections, and substitutes the minimum operation with a mean or a median. Similarly, Bloom Filter follows an approach resembling Count-Min, but always works with Boolean values. Some other examples of widely used sketches include HyperLogLog \cite{flajolet2007hyperloglog}, AMS Sketch \cite{alon1999space}, and Tensor Sketch \cite{pagh2013compressed}.  Recent works have tried to incorporate supervised learning into frequency estimation pipelines \cite{hsu2019learning, aamand2019learned}. Moreover, the work in \cite{kraska2018case} makes a clear link between indexing functions and learnable models showing that several data structures can be learned, including Bloom Filters.

\paragraph{Sketches for learning-based numerical linear algebra}

Recent works on sketching-based numerical linear algebra, have explored learning the sketching projections $S$ \cite{indyk2019learning, indyk2021few, liu2020learning, liu2022tensor}. While these approaches share some similarities to our proposed approach, they have some key distinctions: only linear projections are learned, and the techniques are only applicable to settings that can be framed as matrix decomposition or similar, while our proposed framework is applicable to any learning task as long as a differentiable function can be defined (e.g. predicting weights of an autoencoder).

\paragraph{Multiplexing Networks}

Recently, DataMUX \cite{murahari2022datamux}, a multiplexing approach with neural networks has been proposed, where sequences of samples are compressed into a compact representation that preserves its ordering. While architecturally similar to our approach, DataMux differs conceptually with our proposed approach. DataMux performs an average of multiple embeddings and then disentangles (demultiplexing) each representation to predict a desired label for each element in the sequence. This compact embedding allows for the transmission of utility-preserving information along a channel. On the other hand, the proposed Sketch-Query Network learns a sample permutation invariant compact representation that captures the agglomerate essential information from a set to perform a specific task. In other words, SQNet learns to summarize a set of samples to represent the set as a whole.

\paragraph{Neural Networks and Sketching}

Multiple recent works have attempted to combine the benefits of sketching into multiple aspects of neural networks and machine learning. The work in \cite{daniely2017short} shows that the first layer of a neural network can be replaced by a sketching mechanism with a bounded loss of information. SketchML \cite{jiang2018sketchml} is a framework that makes use of sketches to compress gradient information in a distributed training setting, obtaining considerable speedups on gradient descent-based methods. The work in \cite{ghazi2019recursive} makes use of recursive randomized sketches applied to the intermediate outputs of modular neural networks creating semantic summaries of the data. DiffSketch \cite{li2019privacy} is a framework that makes use of Count-Sketch and its inherent differential privacy properties to perform private distributed training. The work in \cite{wang2020random} applies sketching to reduce the computational requirements for training a ReLU neural network.

\paragraph{Discussion.} While the previously described fields try to obtain a compact representation from data to extract information, there are some key differences, specifically regarding the nature of the computed sketch. A main difference between the sketching operations used for Low-rank approximation (LRA) in numerical linear algebra (NLA) and the sketching approaches applied in data streaming and compressive learning (CL) is that in LRA-based applications it is common to provide a representation with a size that scales with $N$ and allows approximately reconstructing the original data, namely the sketch of a $N \times d$ matrix has size $l \times d$, with $l \ll N$. In many data streaming applications and compressive learning applications, the size of the sketch is constant and independent of $N$, where the complete $N \times d$ dataset is compacted into a representation that captures global information and statistics of interest. Furthermore, data streaming sketching-based applications typically deal with very large dimensionality $d$ inputs (e.g. from thousands to millions), while sketching for LRA deals with medium dimensionality (e.g. thousands), and compressive learning is typically applied to lower dimensional inputs (e.g. from 2 to hundreds). Another differentiating aspect, is the dimensionality of the sketch with respect to the input dimensionality, while in streaming applications it is common that $m \ll d$, in compressive learning we usually have the opposite relationship $m > d$. Furthermore, in many LRA and streaming applications, linear projections are commonly used, while in CL, non-linear mappings are applied to each sample in order to compute the sketch.

\section{Generalization Bounds Proof} 

Stability-based bounds can be easily applied within the Compressive Meta-Learning framework in order to obtain generalization guarantees of the parameters predicted by the Sketch-Query Network. The generalization properties of the predicted parameters $\hat{\theta} = (\psi \circ \Phi)_{\omega}(\mathbf{x})$ are characterized by the complexity of the Sketch and Query networks. Namely, the difference between the empirical $\mathcal{L}$ and expected error $\mathcal{L}^T$ of the predicted parameters can be bounded by the maximum norm of the sketch $\beta_\phi$ and loss function $M_\ell$,  the Lipschitz constant of the Query Network $\rho_\psi$ and of the loss $\rho_\ell$. With probability $1-\delta$:

\begin{equation}
\mathcal{L}^T(\hat{\theta}) < \mathcal{L}(\hat{\theta}) + \frac{2\beta_\phi \rho_\psi \rho_\ell}{N} + (4\beta_\phi \rho_\psi \rho_\ell + M_\ell)\sqrt{\frac{\ln{\nicefrac{1}{\delta}}}{2N}}
\end{equation}

Theorem 12 of \cite{bousquet2002stability} states that the generalization bound of a learning algorithm $A(\mathcal{S})$ with uniform stability $\beta$, a loss function bounded by $M$, and a training set $\mathcal{S}$, with $|\mathcal{S}|=N$, one has that with probability at least $1-\delta$:

\begin{equation}
\mathcal{L}^T(\hat{\theta}) < \mathcal{L}(\hat{\theta}) + 2\beta 
+ (4N\beta + M)\sqrt{\frac{\ln{\nicefrac{1}{\delta}}}{2N}}
\end{equation}

By noting that the uniform stability constant $\beta$ is a function of the maximum norm of the sketch $\beta_\phi$  and  the Lipschitz constant of the Query Network $\rho_\psi$ and of the loss $\rho_\ell$, proving the generalization bound for Sketch-Query Network can be easily proved.

An algorithm $A$ has uniform stability \cite{bousquet2002stability}, with respect to a loss function $\ell$ if for all possible training datasets $\mathcal{S}$:

\begin{equation}
\forall \mathcal{S}, \forall i, \;\; ||\ell(A(\mathcal{S}),\cdot) - \ell(A(\mathcal{S}^{\backslash i}),\cdot)||_{\infty} \leq \beta
\end{equation}

Note that the maximum norm of the sketch is related to how much a sketch can change if the $i$th sample is changed:

\begin{align}
||\Phi(\mathcal{S}) - \Phi(\mathcal{S}^{i})|| &= || \frac{1}{N}\sum_{x_j \in \mathcal{S}}\phi(x_j)  -  \frac{1}{N}\sum_{y_j \in \mathcal{S}^{i}}\phi(y_j)|| \\
&= ||\frac{1}{N} \phi(x_i)  -  \frac{1}{N}\phi(y_i)|| \\
&\leq \frac{2}{N} || \phi(x_i) || \\
&= \frac{2\beta_\phi}{N}
\end{align}

By considering an $\rho_\psi$-Lipschitz Query Network, and a $\rho_\ell$-Lipschitz loss function, it follows that 

\begin{align}
\;\; ||\ell((\psi \circ \Phi)(\mathcal{S}),\cdot) - \ell((\psi \circ \Phi)(\mathcal{S}^{\backslash i}),\cdot)||_{\infty} & \leq \nonumber \beta = \frac{\beta_\phi \rho_\psi \rho_\ell }{N}
\\
\forall \mathcal{S}, \forall i, 
\end{align}

By replacing the uniform stability constant $\beta$ with the stability bounds, we conclude the proof of the proposed bound.

\section{Differential Private Sketch-based Learning}

Because only access to the sketch is needed in CL, and not to the original data samples, sketching-based learning is a good approach when data cannot be shared due to its privacy restrictions. Previous works \cite{chatalic2022compressive} have explored incorporating approximate differential privacy (DP) into the sketch generation in order to provide statistical guarantees about the privacy of the samples used to compute the sketch. Here, we explore the use of 
($\epsilon$, $\delta$)-DP within our proposed Sketch-Query Network.
First, we perform norm clipping at the output of the sketch projection:

\begin{equation}
\phi'(x_i) = \phi(x_i) \min(1,\frac{S}{||\phi(x_i)||})
\end{equation}

where $S = \max_{x}(||\phi(x)||)$ is the maximum L2 norm of the sketch projection across the meta-training samples. $S$ can either be set manually if an upper bound of the value is known, or its empirical maximum across all the samples within the meta-training set can be used. This ensures that when doing inference in new unseen datasets, the L2 norm of the projection will always be bounded for all $x$: $||\phi'(x)|| \leq S$. $S$ also represents the sensitivity of the unnormalized sketch $v = \sum_{i=1}^{N}\phi'(x_i)$. Namely, when a new $x_i$ is added or removed from $v$, its L2 norm will change at most $S \geq ||\sum_{i=1}^{N}\phi'(x_i) - \sum_{i=1}^{N-1}\phi'(x_i)||$. Note that the value of $S$ is computed using only the (meta-)training samples, therefore not violating the privacy requirements when used during test inference. 
We privatize the clipped sum with a Gaussian mechanism parameterized by ($\epsilon_1$,$\delta$), and (optionally) the count with a Laplace mechanism with parameter $\epsilon_2$, composing to a total privacy budget $\epsilon=\epsilon_1+\epsilon_2$. 
The ($\epsilon$, $\delta$)-DP sketch $z_{\epsilon,\delta}$ can be computed as:

\begin{equation}
\label{eq:sup:dp}
z_{\epsilon,\delta} = \frac{v + \xi}{N + \zeta}
\end{equation}

where $\xi \in \R^m$, $\xi \sim \mathcal{N}(\mathbf{0},\,\sigma^{2}\mathbf{I}_m)$ is an additive multivariate Gaussian $m$-dimensional noise with a standard deviation $\sigma$ applied to the unnormalized sketch $v$. The standard deviation is a function of $\epsilon$, $\delta$, and the sensitivity of the sketch $S$, such that $\sigma = \eta(S, \epsilon_{1}, \delta)$, as defined in \cite{chatalic2022compressive, balle2018improving}. $\zeta \sim \mathrm{Laplace}(1 / \epsilon_2)$ is a Laplacian noise applied to the count of elements, $N$, within the sketch, with $\epsilon = \epsilon_1 + \epsilon_2$. Small values of $\epsilon$ and $\delta$ will lead to large values of variance to provide strong privacy guarantees.

DP ensures that any post-processing applied to $z_{\epsilon,\delta}$ will preserve its privacy guarantees. We perform a clamping of the sketch to remove potential out-of-range values due to the noise addition: $z^{c}_{\epsilon,\delta} = \min(\max(z_{\epsilon,\delta}, z_\text{min}), z_\text{max})$, where $\min(\cdot,\cdot)$ and $\max(\cdot,\cdot)$ are applied elementwise, and $z_\text{min}$ and $z_\text{max}$ are the minimum and maximum values of the sketch, either specified manually or by selecting the empirical value during the meta-training process. Then, we can use the query network to learn differentially private parameters: $\hat{\theta}_{\epsilon,\delta} = \psi(z^{c}_{\epsilon,\delta})$

Both the dimensionality of the sketch and the number of samples used to compute it will have an important impact on how much information is preserved after adding differential privacy. For example, for projections $\phi(x_i)$ with an absolute value bounded by 1, their sensitivity is equal to the square root of its dimensionality $S=\sqrt{m}$. Therefore, as more dimensions the sketch has, the more amount of Gaussian noise needs to be added. Furthermore, by looking at Eq. \ref{eq:sup:dp}, one can easily show that:

\begin{equation}
\lim_{N\to\infty} z_{\epsilon,\delta} = z
\end{equation}

\section{PCA and Ridge Regression}
\label{appendix-pca}

\begin{figure*}[!h]
  \centering
  \includegraphics[width=0.85\linewidth]{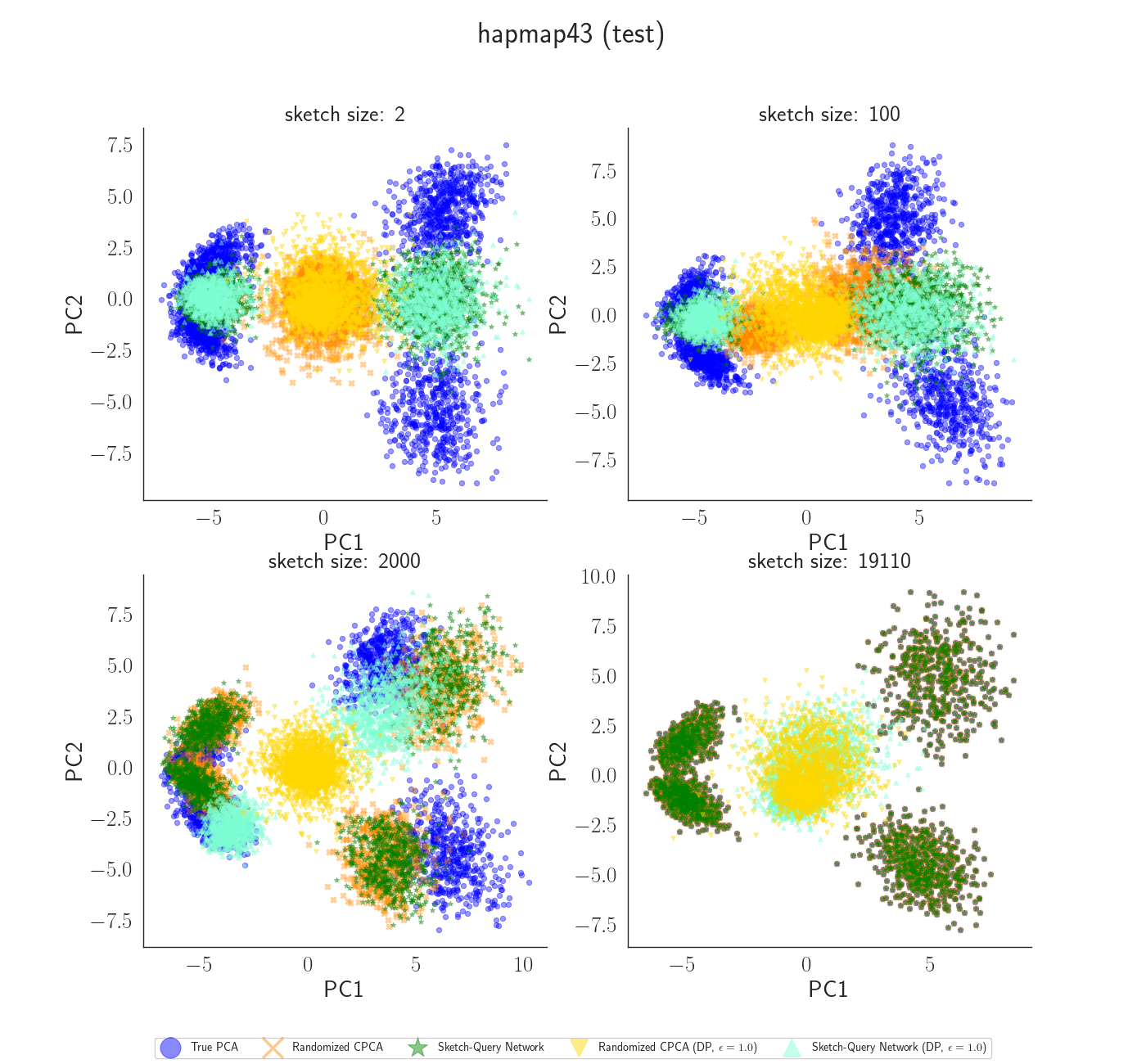}
  \caption{PCA results for the hapmap$_{43}$ dataset with differential privacy results.}
  \Description{Results on PCA for hapmap dataset with differential privacy}
  \label{dp-pca}
\end{figure*}

\begin{figure*}[!h]
  \centering
  \includegraphics[width=0.85\linewidth]{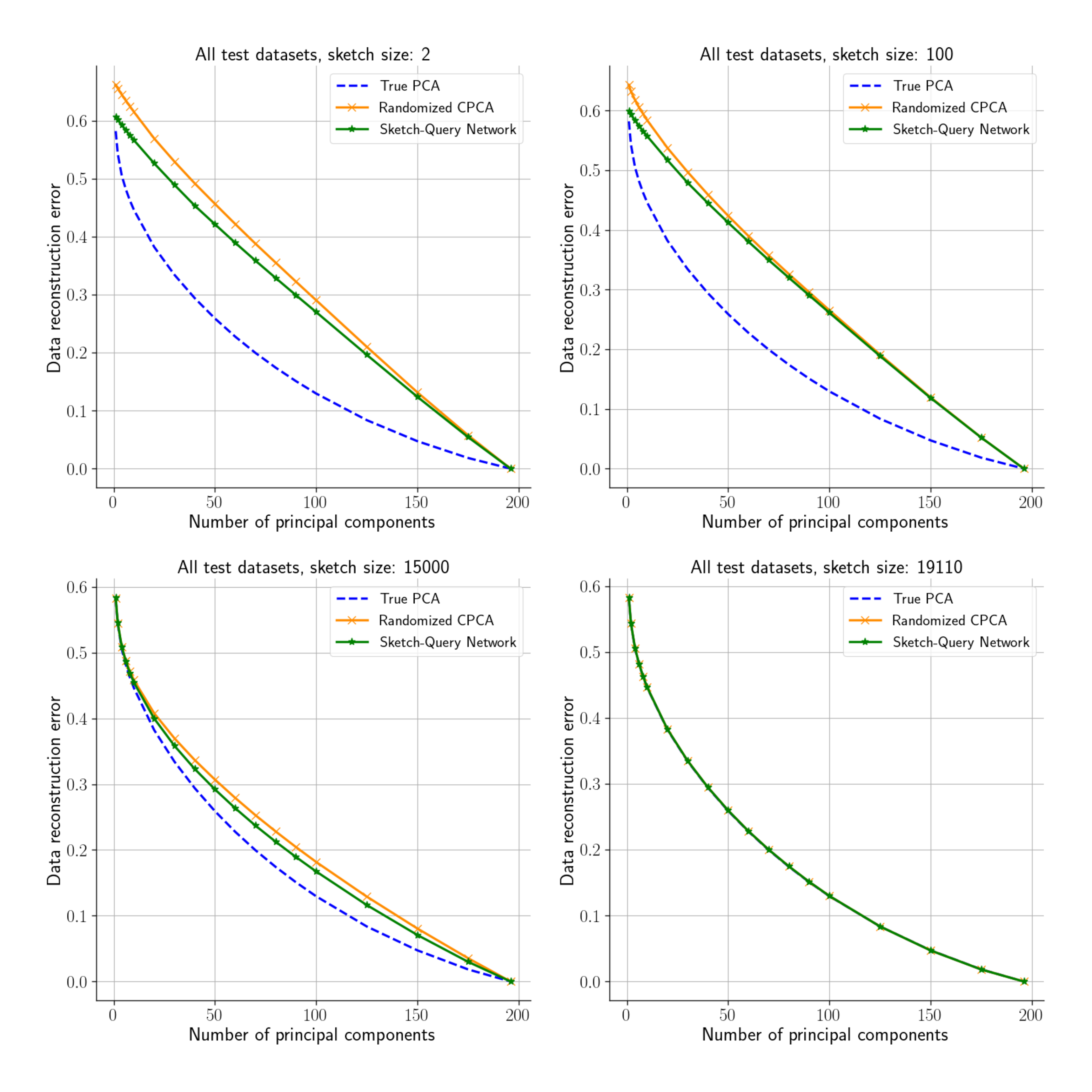}
  \vspace{-20pt}
  \caption{PCA data reconstruction error for different number of principal components considered, and different sketch sizes.}
  \Description{Results on PCA for different sizes}
  \label{fig:curves-barrido}
\end{figure*}

\begin{figure*}[htpb]
  \centering
  \begin{subfigure}[t]{0.48\textwidth}
    \centering
    \includegraphics[width=\linewidth]{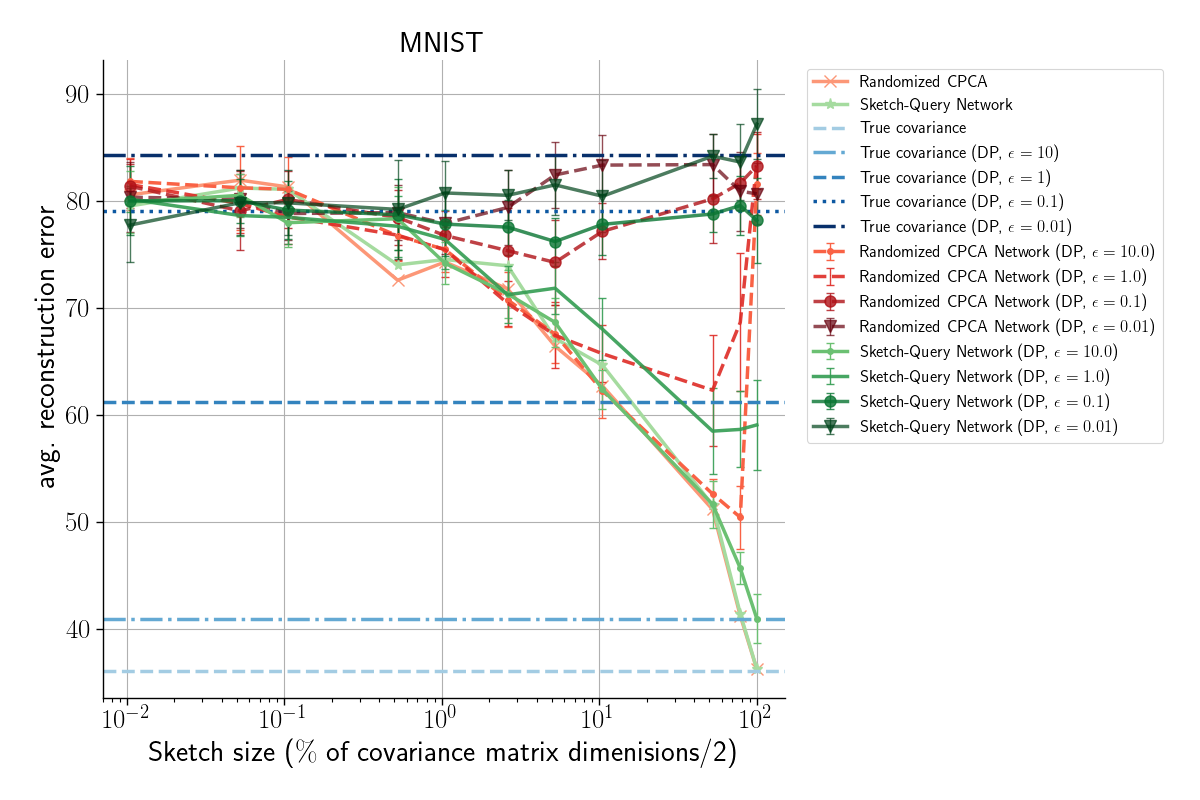}
    \caption{Differential privacy results, for different $\epsilon$.}
    \Description{Results on differential privacy}
    \label{dp-curves}
  \end{subfigure}
  \hfill
  \begin{subfigure}[t]{0.48\textwidth}
    \centering
    \includegraphics[width=\linewidth]{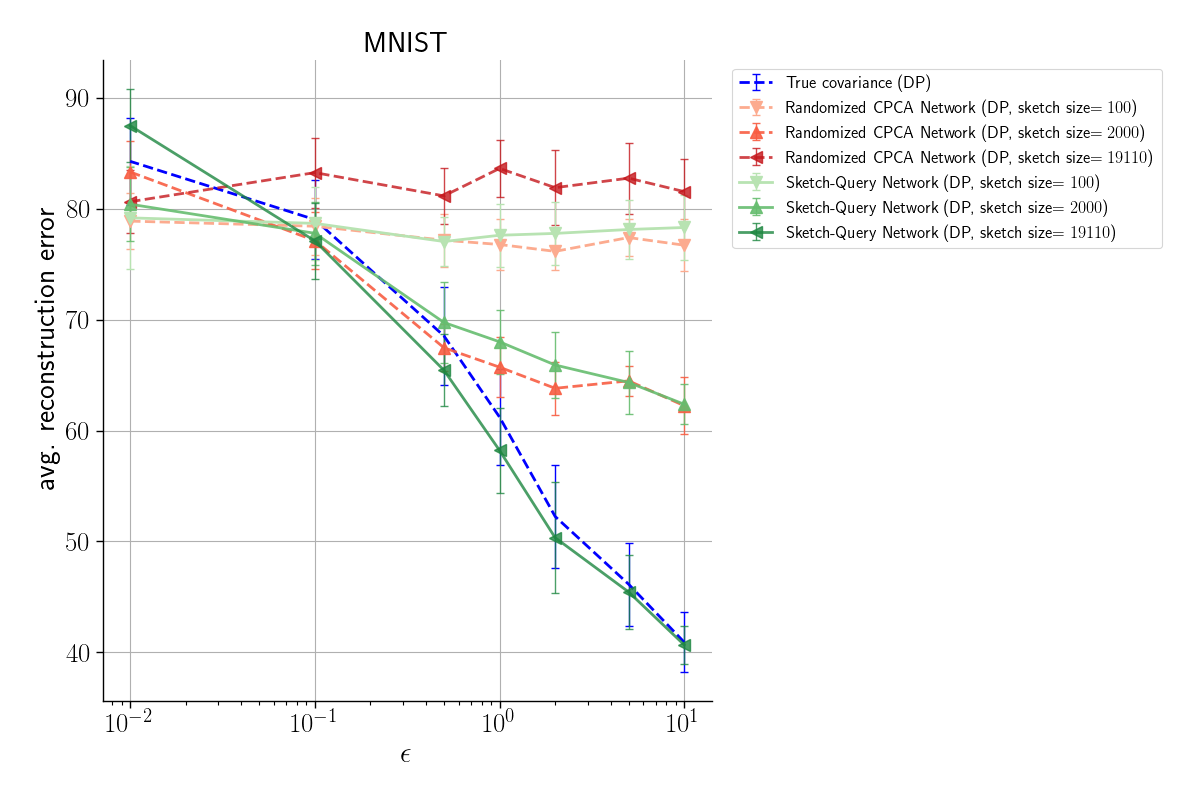}
    \caption{Differential privacy results, for different sketch sizes.}
    \Description{Results on differential privacy}
    \label{dp-curves-2}
  \end{subfigure}
  \vspace{-5pt}
  \caption{Differential privacy results on MNIST. (a) Varying $\epsilon$. (b) Varying sketch size.}
  \label{dp-curves-both}
\end{figure*}

\subsection{Compressive PCA and Linear Regression}

Compressive PCA (CPCA) and Compressive Ridge Regression (CRR) re-frame the task of parameter prediction (Principal Component projection, and linear regression weights/coefficients respectively), with the task of predicting the covariance matrix of the data:

\begin{equation}
\hat{R} = \argmin_{R} \mathbf{C}(R | z)
\label{eq:sup:comp-pca-loss}
\end{equation}

such that $\hat{R} \approx R$, and $R = \frac{1}{N}\sum_{i=1}^{N} x_i x_i^T$. For a given $R$, the ridge regression and PCA parameters can be recovered. Specifically, ridge regression parameters can be obtained as:

\begin{equation}
\label{eq:sup:ridge-reg}
\theta_{\text{Reg}} = R_{12}(R_{22} + \lambda I)^{-1}
\end{equation}

where $R = \begin{pmatrix}
R_{11} & R_{12}\\
R_{21} & R_{22}
\end{pmatrix}$,
$x_i = [x^{(y)}_i, x^{(x)}_i]$ are the regression labels $x^{(y)}_i$ and input features $x^{(x)}_i$ of the $i$th sample concatenated, $R_{12} = R_{21}^T$ is the empirical cross-correlation between labels and features   $R_{12} = \frac{1}{N}\sum_{i=1}^{N} x^{(y)}_i x^{(x)T}_i$, and $R_{22}$ is the correlation between features $R_{22} = \frac{1}{N}\sum_{i=1}^{N} x^{(x)}_i x^{(x)T}_i$. $\lambda$ is the regularization parameters which is typically set through cross-validation and in our experiments is set as $\lambda = ||R_{22}||_F$ for simplicity. 

The PCA parameters can be recovered as the eigendecomposition of the covariance matrix:

\begin{equation}
\label{eq:sup:eigenpca}
R = \theta D \theta^T
\end{equation}

where $\theta$ are the eigenvectors and $D$ is a diagonal matrix with eigenvalues of $R$. Both Eq. \ref{eq:sup:ridge-reg} and \ref{eq:sup:eigenpca} show that $R$ is a sufficient statistic to properly recover the ridge regression and PCA parameters.

By simply setting $\phi(x_i) = x_ix_i^T$ a sketch such as $z = R$ provides a perfect recovery from the parameters of interest. However, the size of $R$ grows quadratically with the number of input dimensions, therefore Compressive Learning adopts a more compact representation with dimensionality $m \ll d^2$:  

\begin{equation}
z = \frac{1}{N}\sum_{i=1}^{N}A\text{vec}(x_i x_i^T) =A\text{vec}(R)
\label{eq:comp-pca-sketch-2}
\end{equation}

where $\text{vec}(\cdot)$ flattens the $d \times d$ matrix into a $d^2$ vector, and $A$ is a random matrix with dimensions $m \times d^2$. Note that $\phi(x_i) = A\text{vec}(x_i x_i^T)$ can be equivalently framed as random square features $\phi(x_i) = (Bx_i)^2$, where $B$ can be obtained from a given $A$. Compressive PCA and Ridge Regression \cite{gribonval2020sketching, gribonval2021compressive} re-frames the learning problem as minimizing the following objective: 

\begin{equation}
\hat{R} = \argmin_{R} ||A\text{vec}(R) - z ||^2
\label{eq:comp-pca-loss-2}
\end{equation}

where $A$ is the given random matrix and $z$ is the empirical sketch, computed as described in Eq. \ref{eq:comp-pca-sketch-2}. $\hat{R}$ can be found by minimizing Eq. \ref{eq:comp-pca-loss-2} through any desired optimization procedure. In fact, the linear projection that minimizes Eq. \ref{eq:comp-pca-loss-2} can be obtained in closed-form by computing the pseudo-inverse of the randomized projection: 

\begin{equation}
\text{vec}(\hat{R}) = A^+z
\end{equation}

Because $R$ is a symmetric matrix, the sketching process (Eq. \ref{eq:comp-pca-sketch-2}) and optimization objectives (Eq. \ref{eq:comp-pca-loss-2}) can be framed by using only the vectorized lower (or upper) triangular elements of $R$, i.e. replacing $\text{vec}(\hat{R})$ by $\text{vec}_{\text{LT}}(\hat{R}) \in  \R^{\frac{d(d+1)}{2}}$. 

\subsection{Neural CPCA and CRR}

Here we provide extended details for the Neural-based CPCA and CRR applications described in \Cref{ssec:pca}. First, note that we minimize the L1 error in Eq. \ref{eq:comp-pca-neural-loss}, but other distance metrics such as the L2 loss could be used instead of the L1. We select the L1 loss as it provided a more stable training process.

We standardize all our meta-training and evaluation datasets to have zero mean and unit variance, to ensure that the ground truth covariance matrix has bounded values, such that $|\text{vec}_{\text{LT}}(R)_j| \leq 1$. Note that in practical applications, the information of the mean and variance can be treated as additional sketches. By using a Tanh activation function at the end of the query network, we ensure that the predicted covariance matrix has the same range of values as the ground truth covariance matrix. After meta-training the sketch-query network, the covariance matrix $\text{vec}_{\text{LT}}(\hat{R})$ can be predicted in new unseen dataset, and the PCA or regression coefficients can be easily recovered.

\subsection{Experimental details}

In order to evaluate the quality of the predicted PC projection matrix $\hat{\theta}$, we compute the reconstruction error on the testing datasets as:

\begin{equation}
\text{Err}_{\text{PCA}}(X_{\text{test}},\hat{\theta}) =  \frac{1}{d}\sum_{r=1}^{d} ||X_{\text{test}} - \hat{\theta}_r\hat{\theta}_r^TX_{\text{test}}||^2
\label{eq:comp-pca-eval-metric}
\end{equation}

where $\hat{\theta}_r$ is the PCA projection matrix containing only the first $r$ principal components. The error averages the reconstruction through the values of $r$ from $1$ to $d$. We report the relative errors, by computing the logarithmic ratio between the average reconstruction error (LRE) of PCA with the predicted PC projection $\hat{\theta}$, $\text{Err}_{\text{PCA}}(X_{\text{test}},\hat{\theta})$, and the average reconstruction error of PCA with the ground-truth PC projection computed from the empirical covariance matrix using the complete dataset, $\theta$, $\text{Err}_{\text{PCA}}(X_{\text{test}},\theta)$:

\begin{equation}
\text{LRE}_{\text{PCA}}(X_{\text{test}},\hat{\theta}, \theta) =  \log \frac{\text{Err}_{\text{PCA}}(X_{\text{test}},\hat{\theta})}{\text{Err}_{\text{PCA}}(X_{\text{test}},\theta)}
\label{eq:comp-pca-lre-eval-metric}
\end{equation}

Similarly, we evaluate the ridge regression error by computing the mean square error as follows:

\begin{equation}
\text{Err}_{\text{Reg}}(X_{\text{test}},\hat{\theta}) =  ||X^{(y)}_{\text{test}} - \hat{\theta}X^{(x)}_{\text{test}}||^2
\label{eq:comp-regression-eval-metric}
\end{equation}

where $\hat{\theta}$ are the regression coefficients, $X^{(x)}_{\text{test}}$ and $X^{(y)}_{\text{test}}$ are the input features and continuous labels for a given dataset. Similarly to the task of PCA, we compute the log ratio of the error when the predicted regression coefficients are used relative to the error when using the ground truth regression coefficients obtained by using the empirical covariance matrix:

\begin{equation}
\text{LRE}_{\text{Reg}}(X_{\text{test}},\hat{\theta}, \theta) =  \log \frac{\text{Err}_{\text{Reg}}(X_{\text{test}},\hat{\theta})}{\text{Err}_{\text{Reg}}(X_{\text{test}},\theta)}
\label{eq:comp-regression-lre-eval-metric}
\end{equation}

The Sketch-Query-Network for Compressive PCA and Compressive Ridge regression is trained by computing sketches with 4096 samples randomly selected per dataset, with a total of 64 randomly chosen meta-training datasets per batch. A total of 196 input features are randomly selected at each iteration during training. The Adam optimizer with a learning rate of $3 \times 10^{-5}$ is used.

\subsubsection{Extended CPCA Results}

Figure~\ref{dp-pca} shows a visualization on the first two principal components for a human genome dataset, for different sketch sizes and differential privacy. Projections with the Sketch-Query Network follow the ground truth better than with Randomized CPCA, both with added differential privacy (shown at $\epsilon=1.0$) and without. In Figure~\ref{fig:curves-barrido} we show the data reconstruction error for the test datasets when different number of principal components are used, for different sketch sizes. Figure~\ref{dp-curves} and Figure~\ref{dp-curves-2} show results on the MNIST dataset including differential privacy for different $\epsilon$ and different sketch sizes, respectively. 
We provide tables with information on how the datasets are organized in  \Cref{resultstable1} and \Cref{resultstable2}, and a per-dataset break-down of the reconstruction error in Table~\ref{resultstable3_part1} and Table~\ref{resultstable3_part2}. 

After meta-training, the Sketch-Query-Network can predict the PCA projection and the linear regression parameters for a given dataset in an average time of 0.1 seconds in a V100 GPU.

\section{Compressive k-means}

\begin{figure*}[htpb]
  \centering
  \includegraphics[width=0.8\linewidth]{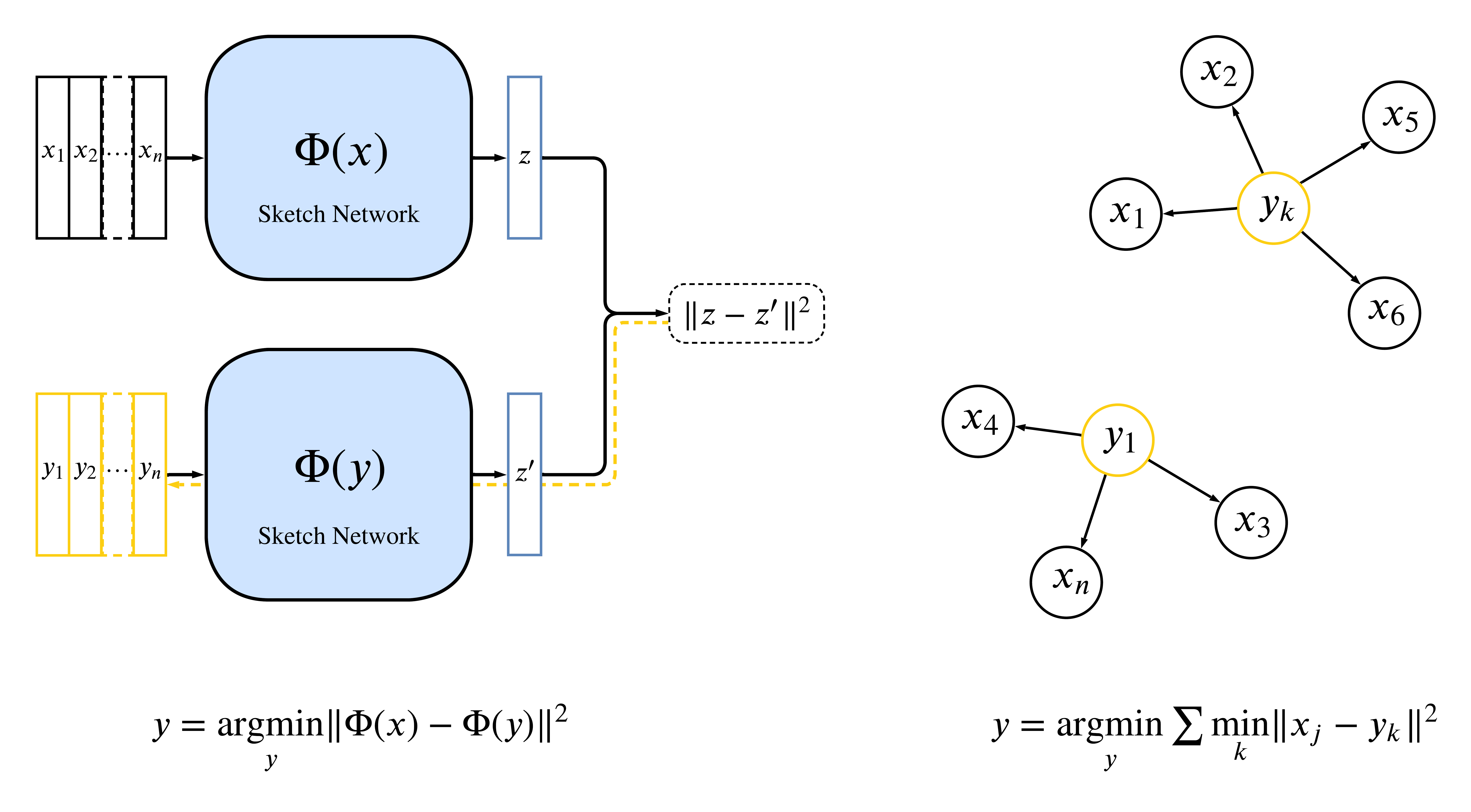}
  \vspace{-10pt}
  \caption{Compressive vs traditional k-means. Dashed yellow line (left) represents the gradient.}
  \Description{Comparison of compressive vs. traditional k-means}
  \label{fig:kmeans_comp}
\end{figure*}

\begin{figure*}[htpb]
  \centering
  \includegraphics[width=1\linewidth]{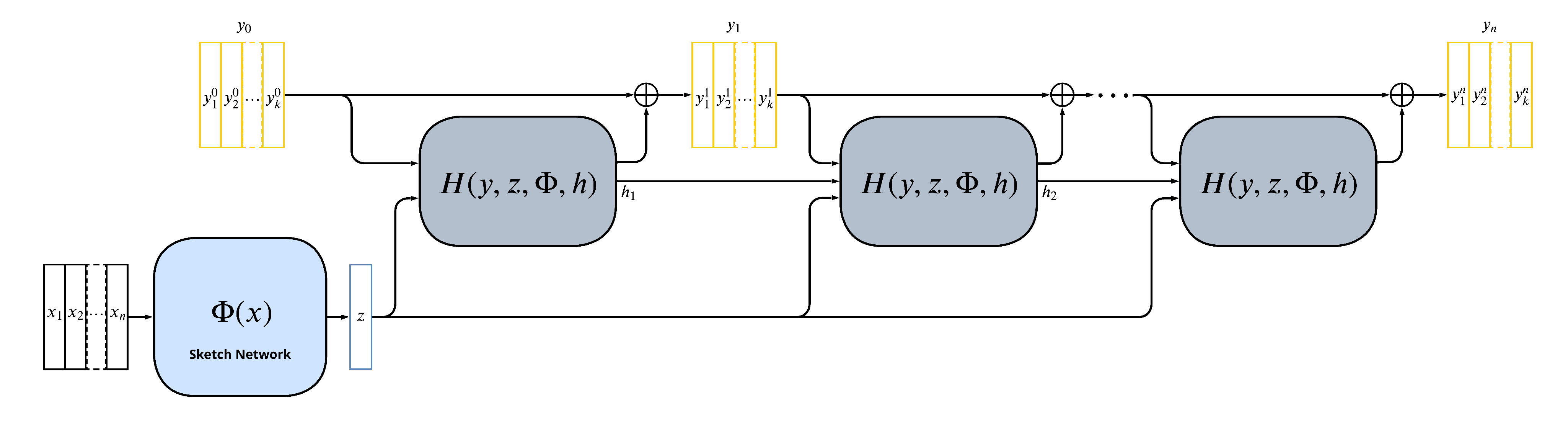}
  \caption{Unrolled compressive k-means with Adam-based SQNet.}
  \Description{Diagram of unrolled compressive k-means}
  \label{fig:unrolledkmeans}
\end{figure*}

Compressive k-means (Figure~\ref{fig:kmeans_comp}) tries to ``learn'' (infer) the $k$ centroids reducing the following error function:

\begin{equation}
\tilde{\mathbf{y}} = \argmin_{\mathbf{y}} || \Phi(\mathbf{y}) - \Phi(\mathbf{x}) ||^2 = \argmin_{\mathbf{y}} || \Phi(\mathbf{y}) - z ||^2
\label{eq:comp-kmeans2}
\end{equation}

Different optimization approaches can be adopted to find the optimal centroids $\mathbf{y}$. Here we use gradient-based optimization and unroll (unfold) the optimization procedure to be able to learn the projection function, see Figure~\ref{fig:unrolledkmeans}. We provide a detailed description of the unfolded process with SGD for simplicity, but we adopt the Adam optimizer \cite{kingma2014adam} in our experiments, which, after unfolding, leads to the following iterative process:

\begin{equation}
\mathbf{\hat{y}_i} = \psi(z,\mathbf{\hat{y}_{i-1}}, \Phi) = \mathbf{\hat{y}_{i-1}} - H( \mathbf{\hat{y}_{i-1}}, z, \Phi, h_{i})
\label{eq:adam_unrolled_0}
\end{equation}

Here $H$ is a function of the empirical sketch $z$, the current centroid estimates $\mathbf{\hat{y}_{i-1}}$, and a memory $h_{i-1}$:

\begin{equation}
h_i = (m_i, v_i) = (\beta_1 m_{i-1} + (1-\beta_1) g_i, \beta_2 v_{i-1} + (1-\beta_2) g_i^2)
\label{eq:adam_unrolled_1}
\end{equation}

\begin{equation}
g_i = g(\mathbf{\hat{y}_{i-1}},z) = \nabla \mathcal{L}(\Phi(\mathbf{\hat{y}_{i-1}}),z)  = \nabla ||\Phi(\mathbf{\hat{y}_{i-1}})-z||^2
\label{eq:adam_unrolled_2}
\end{equation}

\begin{equation}
H( \mathbf{\hat{y}_{i-1}}, z, \Phi, h_{i-1}) = \frac{\alpha}{1-\beta_1^i} \frac{m_i}{\sqrt{\nicefrac{v_i}{1-\beta_2^i}}}
\label{eq:adam_unrolled_3}
\end{equation}

The previous equations simply describe the Adam optimization steps. Note that the centroid inference procedure depends both on the optimization algorithm used (e.g. Adam) and the sketching function $\Phi$, and by unfolding the iterative inference of the centroids both the sketching function and the optimization procedure can be jointly learnt such that the loss function of traditional k-means is minimized. The training procedure is as follows: (1) compute the sketch $\Phi(\mathbf{x})$ with the set $\mathbf{x}$, (2) randomly instantiate initial centroid cluster estimates $\mathbf{\hat{y}_0}$, (3) iteratively update $\mathbf{\hat{y}_i}$ following the equations \ref{eq:adam_unrolled_0},\ref{eq:adam_unrolled_1},\ref{eq:adam_unrolled_2},\ref{eq:adam_unrolled_3}, (4) after $t$ steps, compute the k-mean loss (Eq. \ref{eq:comp-kmeans2}) or the Hungarian loss with
$\mathbf{x}$ and $\mathbf{y_t}$, (5) finally, update the parameters of $\Phi$ to reduce the loss.

\section{Sketch-Conditional Autoencoders}

\begin{figure*}[!htpb]
  \centering
  \includegraphics[width=0.9\linewidth]{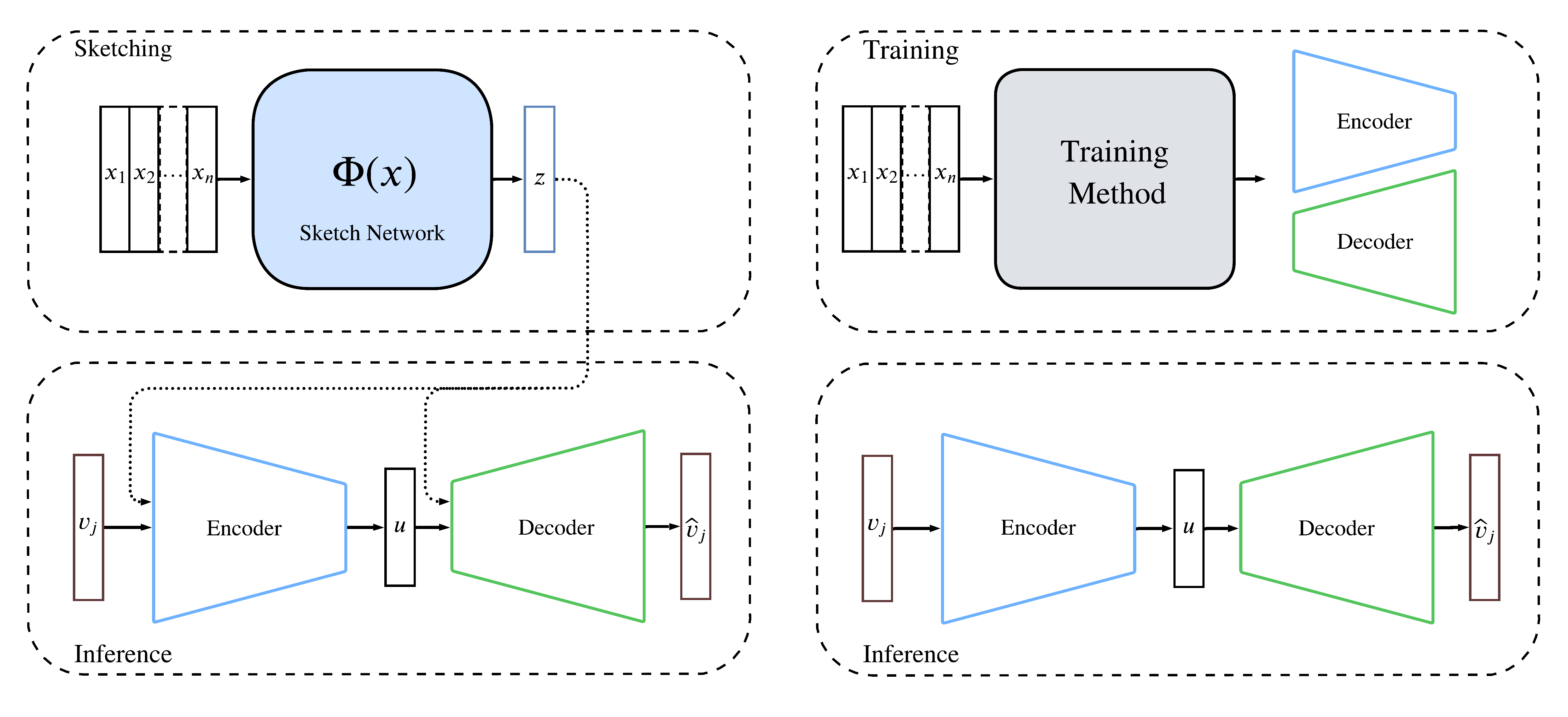}
  \caption{Comparison of traditional autoencoders and the sketch-conditional autoencoder.}
  \Description{Comparison of traditional autoencoders with sketch-conditional autoencoder}
  \label{fig:sketchae}
\end{figure*}

The sketch-based autoencoder introduced in \cref{ssec:autoencoder} makes use of a sketch-conditional encoder $u = f_{\theta}(x)$ and a sketch-conditional decoder $\hat{x} = g_{\theta}(u)$. 
Figure~\ref{fig:sketchae} depicts the differences between regular autoencoders and sketch-conditional autoencoders. Regular autoencoders are trained with a dataset of interest following the same underlying distribution as the samples expected during inference. On the other hand, sketch-based autoencoders substitute the traditional training stage with a fast sketching operation (i.e. inference with the sketch network) to adapt a meta-trained autoencoder to a new dataset or target distribution.
An important aspect of the sketch-based autoencoder is its built-in active learning nature. Because dataset-specific learning is substituted by sketching, by simply updating the sketch, the network can adapt to new datasets, to distribution shifts, or to improve its performance by including more samples in the sketch. As linear sketching is used, adding or removing a sample from the sketch can be easily done with an addition or subtraction operation.

Another way to frame sketch-conditional autoencoders is by understanding the sketching network as a hypernetwork \cite{ha2017hypernetworks} that predicts some weights of a primary network (encoder-decoder pair). While common hypernetworks generate a set of weights for a given input sample, here the weights are generated given a complete dataset. 
Furthermore, sketches can be seen as external memories for memory-augmented neural networks, which can be easily updated (by simply merging sketches), and can capture in a distributed manner the properties and shape of the density distribution.

\subsection{Compression and information theory}

The presented autoencoders map $d$-dimensional boolean sequences into $m$-dimensional float embeddings. Note that each float is represented by using 32-bits so one could expect to obtain an autoencoder that has a ratio of input dimension over embedding dimension of 32 ($d/m = 32$) with no reconstruction error. However this is not possible due to the nature of floats (not every possible 32-bit sequence maps to a numeric float value). Furthermore, the implicit smoothness of neural networks and the learning algorithms limit which compression functions can actually be learnt. In fact, in this work we explore networks with an embedding dimension of $50$ ($d/m = 1000/50 = 20$) leading to non-zero reconstruction errors.
The role of the sketch within an autoencoder can be framed within an information theory perspective. Traditional lossless compression techniques such as Shannon coding, make use of the underlying probability distribution of the data to assign fewer bits to the frequent subsequences, and more bits to the more rare subsequences. However, if all elements are equally likely to appear, or the underlying distribution is not knowable or completely random, the best that can be done is to assign one bit for each boolean value, obtaining no compression gain. In a similar manner, the sketch acts as a hidden representation of the underlying distribution of the data, which is used by the encoder and decoder to provide a more compact encoded representation of new samples, or similarly, obtaining lower reconstruction errors.

\subsection{Experimental Details}
We train three main network configurations: (a) a regular autoencoder (AE), (b) a mean conditional autoencoder (+M) and (c) a mean and sketch conditional autoencoder (+MS). Each of the three networks uses the same base architecture, where the only difference is the additional weights included in the first layer of both the encoder and the decoder for the dynamic biases. For this application, we use for both encoder and decoder a residual LayerNorm GELU-based fully connected network (see \Cref{fig:net_arch}) with 5 hidden layers and each with a hidden dimension of 4096. The sketch network includes three hidden layers with a hidden dimension of 4096.

As a baseline, we train dataset-specific autoencoders with the RAdam \cite{liu2019radam} optimizer, a learning rate of 0.0001, and a batch size of 1024. We compute the validation loss every 1000 weight updates and stop the training process if the loss has not decreased after fifteen evaluations. We train the same architecture (regular autoencoder) with the randomized binary MNIST. For this dataset a learning rate of 0.00001 and a batch size of 4096 is used. The randomized binarized MNIST is also used to train the mean-conditional and sketch+mean-conditional networks. First, the mean-conditional network is trained until the validation loss does not decrease for more than 50 evaluations. A learning rate of 0.00001 is used with a batch size of 1152. The (conditional) mean is computed using 128 different samples and concatenated to each element of the batch. After the mean-conditional autoencoder is trained, the weights are used as initialization of the sketch+mean-conditional autoencoder. The sketch+mean-conditional autoencoder is trained with the same learning rate, a batch size of 1024, and the sketch is computed using 64 samples. The number of samples used to compute the sketch is increased to 128 after 50 evaluations without improvement of the validation loss.

Note that during evaluation, the autoencoder +MSK setting is obtained by simply using the trained sketch+mean-conditional network (+MS) and computing a specific sketch for each class.

\subsection{Analysis of the sketch}

\begin{figure}[htpb]
  \centering
  \includegraphics[width=0.7\linewidth]{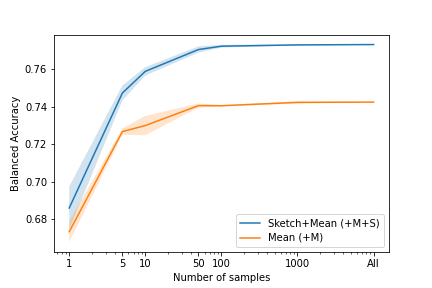}
  \caption{Balanced accuracy of sketch+mean and mean conditional autoencoders with a variable number of samples used to compute the sketch.}
  \Description{Results on bal acc}
  \label{fig:sketch_mean_samples}
\end{figure}

We explore the effect of the sketch resolution on the reconstruction accuracy. Specifically, we run the mean and sketch+mean conditional autoencoders with sketches generated with a variable number of samples from the training set. Figure~\ref{fig:sketch_mean_samples} shows the mean and standard deviation of the balanced accuracy when including from one to all samples of the training set in the sketch. As can be observed, the improvement grows logarithmically, obtaining marginal improvements after more than 1,000 samples have been included.

\section{Data streaming applications}
\label{ssec:Frequency Estimation}

\subsection{Zipf dataset}
Zipf distributed datasets are commonly used to evaluate and characterize the performance of frequency and membership estimation algorithms \cite{cormode2005improved, charikar2002finding, bloom1970space}. The following form is frequently adopted for the marginal distribution of a feature $j$:

\begin{equation}
p_j = \frac{1}{j^{\alpha}}
\label{eq:Zipf-simple}
\end{equation}

Here the features are ordered from most to least frequent with $p_j > p_{j+1}$ (i.e. feature $j$ is more frequent than $j+1$) and $\alpha$ is a parameter that characterizes the level of skewness of the distribution. We propose an extended definition that adds a scaling factor, $\beta$, for simulations with small $\alpha$:

\begin{equation}
p_j = \beta^{\max(1-\alpha,0)} \frac{1}{j^{\alpha}}
\label{eq:Zipf-appendix}
\end{equation}

where $\beta$ provides a scaling factor for samples with $\alpha < 1$.

The first application consists of predicting the frequency (i.e. normalized counts) for each feature within a set of samples. We adopt a setting commonly found in data streaming where the number of elements in a given vocabulary needs to be computed and each element can be represented as a binary vector by using a one-hot encoding. Specifically, the query output we are trying to predict is $y = \frac{1}{N}\sum_{i=1}^{N}x_i$, where $x_i \in \{0,1\}^d$. Furthermore, the vectors are assumed to come from a distribution potentially sparse and skewed. 
In particular, we make use of data where the frequency of each Boolean feature is distributed following a Zipf-like distribution where feature $j$ has a probability $p_j = \beta^{\max(1-\alpha,0)} j^{-\alpha}$
and feature $j$ is equally or more frequent than feature $j+1$. $\alpha$ controls the amount of skewness with $\alpha =0$ leading to a uniform distribution and $\alpha>1$ leading to skewed distributions. $\beta$ is the background probability when $\alpha = 0$ and has no effect for $\alpha > 1$. Note that we treat each feature as independent, and $p_j$ represents the marginal probability for each feature; therefore, the sum of all $p_j$ do not need to add to 1. 

With a sketch size equal to the input dimension ($m = d$) this becomes a trivial task, where applying an identity mapping (i.e. $\phi(x_i) = x_i$) followed by a \textit{mean} provides the exact frequency estimate. Here we focus on the non-trivial scenario where $m < d$ and consider $m = 100$ and $m = 10$ with data of $d = 1000$ dimensions.
We train the SQNet to minimize the binary cross entropy between the predicted set average $\hat{y}$ and the normalized frequency $y$. The query network only takes as input the sketch $z$ and does not make use of any auxiliary inputs.
The networks are trained with randomly generated batches of  binary vectors $x_i$ using the Zipf distribution. During training, the number of elements $n$ within the input set $\mathbf{x}$, and $\alpha$ and $\beta$ are chosen randomly at each batch, with $n$ ranging from 1 to 100, and $\alpha$ and $\beta$ between 0 and 2, and 0 and 1 respectively. Furthermore, the features are randomly permuted at each iteration. This randomized training allows us to obtain a network that can generalize well to a wide variety of datasets with different levels of skewness.

\begin{figure*}[htpb]
  \centering
  \includegraphics[width=0.85\linewidth]{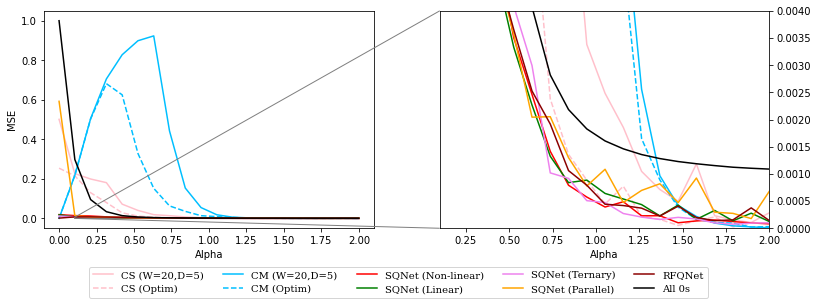}
  \caption{Mean square error (MSE) for frequency estimation with SQNet (Non Linear), SQNet (Linear), SQNet (Ternary), SQNet (Parallel), RFQNet, CM, and CS in a Zipf-distributed dataset for different $\alpha$ values and $\beta = 1.0$ using a sketch size of 100. Right plot shows a zoomed–in view of the MSE range [0,0.0040].}
  \Description{MSE for frequency}
  \label{fig:frequency}
\end{figure*}

We explore using different sketching networks: non-linear MLP networks with skip connections (Non-linear), a linear layer (Linear), and a linear layer limited to ternary values (0,1, and -1) (Ternary) trained following the quantization approach used in \cite{li2016ternary}, and smaller SQNets applied in parallel, each taking only 10\% of the input dimension (Parallel). The query network is always a non-linear residual MLP network. For both Sketch and Query Network we explore multiple architectural variations: using different activation functions, pooling layers, and hidden layer dimensions. The use of non-linear networks is only useful in settings where the input consist of multi-hot encoding. When the input is one-hot encoding, a linear layer is sufficient and is equivalent to learnt embeddings commonly used to encode words from a vocabulary in NLP applications. Sketching performed with ternary linear layers can benefit from efficient low-latency implementations in a similar way as Count-Sketch or Count-Min do. Details of the experimental setting are reported in \Cref{sec:exp_data_streaming applications}.
 
In order to assess the importance of learning a good sketching function, we replace $\Phi(\mathbf{x})$ by a set of random feature projections followed by a mean pooling operation, as a baseline from CL, where only the query network is learnt. We refer to this setting as RFQNet (Random Features + Query Network). Both SQNet and RFQNet are compared with well established frequency estimation algorithms: Count-Min (CM) \cite{cormode2005improved}  and Count-Sketch (CS) \cite{charikar2002finding}. We look for the optimal hyperparameters of CS and CM for a given $\alpha$ and $\beta$. 

Figure~\ref{fig:frequency} shows the mean square error of a Zipf dataset, with $m=100$, for SQNet, RFQNet, and CS and CM when using optimal hyperparameters (Optim), and when using fixed sub-optimal hyperparameters (W=20, D=5). Additionally, we include the performance when predicting always zero (All 0s). SQNet outperforms all competing methods providing a significant increase in accuracy for datasets with $\alpha \leq 1$, and matching CM performance with almost zero error for extremely skewed datasets ($\alpha \geq 1.5$). \Cref{sec:exp_data_streaming applications} provides additional experiments in real world tabular datasets.

The second application, membership estimation, consists of detecting if a feature $j$ is present in at least one sample from a set $\mathbf{x} = \{x_1, x_2, ..., x_n\}$, with $x_i \in \{0,1\}^d$.
Specifically, the query that we aim to predict is:

\begin{equation}
y_j =
\left\{
	\begin{array}{ll}
		1  & \mbox{if } \sum_{i=1}^{N}x_{ij} > 0 \\
		0 & \mbox{if } \sum_{i=1}^{N}x_{ij} = 0
	\end{array}
\right.
\end{equation}

where $x_{ij}$ indicates the feature $j$ at sample $i$, and $y_j$ is the membership indicator of feature $j$. This application can be framed as a binarized version of frequency estimation. In fact, we apply the same training and testing procedure, and compare SQNet with RFQNet and the established method Bloom Filters (BF) \cite{bloom1970space} with both the Zipf-distributed dataset and the previously described tabular datasets. The results follow the same trend as in frequency estimation, with SQNet surpassing competing methods. A more in-depth discussion of the experiments with Zipf and tabular datasets as well as an analysis of the hyperparameters of SQNet, RFQNet, and BF can be found in \Cref{sec:exp_data_streaming applications}.

\subsection{Frequency and membership estimation}

\begin{figure*}[htpb]
  \centering
  \includegraphics[width=0.85\linewidth]{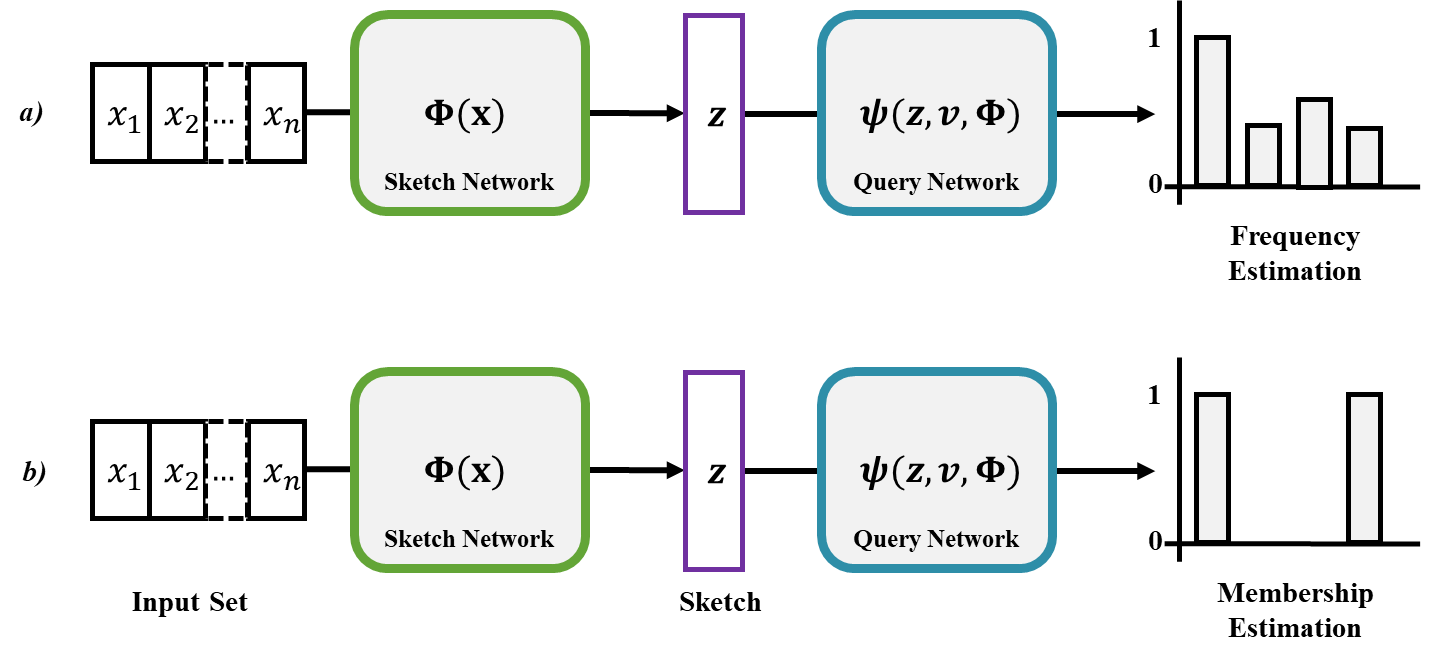}
  \caption{Frequency estimation and membership estimation with SQNet.}
  \Description{Frequency with SQNet}
  \label{fig:freq_memb}
\end{figure*}

Frequency and membership estimation are explored in section \ref{ssec:Frequency Estimation} (Figure~\ref{fig:freq_memb}). We frame frequency estimation as a task of estimating the normalized frequency for each feature given a set of samples $\mathbf{x} = \{x_1, x_2, ..., x_N\}$, with $x_i \in \{0,1\}^d$:

\begin{equation}
y = \frac{1}{N}\sum_{i=1}^{N}x_i
\label{eq:freq-query}
\end{equation}

Here $y$ is a $d$-dimensional vector wherein each dimension $y_j$ represents the normalized frequency for feature $j$. Note that previous works frame frequency estimation without the normalizing factor $N$ \cite{cormode2005improved, charikar2002finding}. Similarly, membership estimation can be framed as predicting a binary indicator per feature describing whether a feature $x_j$ is present or not in a set:

\begin{equation}
y_j =
\left\{
	\begin{array}{ll}
		1  & \mbox{if } \sum_{i=1}^{N}x_{ij} > 0 \\
		0 & \mbox{if } \sum_{i=1}^{N}x_{ij} = 0
	\end{array}
\right.
\end{equation}

Here $y$ is a $d$-dimensional vector where each dimension $y_j$ represents the binary membership indicator for feature $j$. Note that one can binarize the frequency vector to obtain the membership indicator vector. For frequency estimation, the value of $N$ will dictate the resolution of the frequency estimates. For membership estimation, the value of $N$ needs to be properly taken into account as for $p_i > 0$ a large enough $N$ will lead to membership indicators where all values are $1$. Therefore, a proper tuning of membership estimation method's parameters need to include the number of elements in the set.

\subsection{Count-Min, Count-Sketch, and Bloom Filter as neural networks}

Count-Min, Count-Sketch, and Bloom Filter can be framed as Sketch-Query Networks with linear projections with fix weights as sketching functions and a fixed query function. For example, a Count-Sketch with sketch size of $m$ and $N_w=m$ and $N_d=1$, with $N_w$ representing the output dimensionality of the linear projection, and $N_d$ the number of linear projections, can be represented as follows:

\begin{equation}
z_j = Wx_j
\end{equation}

with $z_j$ representing the sketch projection of sample $x_j$, and $W$ is the linear projection matrix with dimensionality $d \times m$ and ternary values $W_{l,k} \in \{-1,0,1\}$. The dataset-level sketch can be obtained by performing a mean (or sum) pooling operation:

\begin{equation}
z =  \frac{1}{N}\sum_{j=1}^{N}z_j
\end{equation}

The estimated frequency vector can be obtained by applying the transposed linear projection:

\begin{equation}
\hat{y} = W^Tz
\end{equation}

Count-Min differs from Count-Sketch by limiting $W_{l,k} \in \{0,1\}$ and using the maximum as pooling layer. Bloom Filter adds the additional constraint of limiting the elements of $z$ to be booleans $z_{j} \in \{0,1\}^m$. If $N_d > 1$, then a total of $N_d$ projections and transposed projections are performed in parallel obtaining a total of $N_d$ estimated frequencies. The multiple sets of estimated frequencies are combined through a mean operation for Count-Sketch and with a min operation for Count-Min and Bloom Filters.

\subsection{Sketching functions and scalability}

We explore different neural network architectures for the Sketching Network. Specifically, we explore non-linear networks, particularly residual MLPs, a linear mapping, and a ternary linear layer. Furthermore, we investigate the effectiveness of employing multiple smaller networks in parallel.

\begin{figure*}[!ht]
  \centering
  \includegraphics[width=0.9\linewidth]{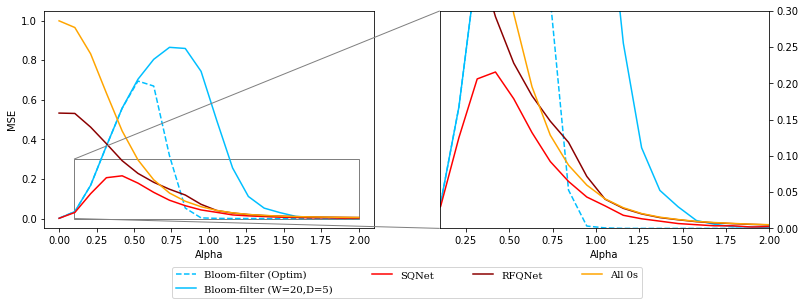}
  \caption{Mean square error (MSE) for membership estimation with SQNet, RFQNet, and BF, in a Zipf-distributed dataset for different $\alpha$ values and $\beta = 0.5$. Right plot shows a zoomed–in view of the MSE range [0,0.3]}
  \Description{MSE for membership estimation}
  \label{fig:membership}
\end{figure*}

\subsubsection{Linear and non-linear sketching networks}
Depending on the application, two different scenarios can be found: (a) the input vector $x_j$ can contain a single one-hot encoded value (i.e. $||x_j|| = 1$) or (b) it can contain multi-hot elements (i.e. $||x_j|| \geq 1$). Non-linear networks can provide improved accuracy in setting (b) as correlations between features can be exploited, however, in setting (a) both linear and non-linear projections can learn the same set of functions, providing no benefit using non-linear mappings over linear ones.

Real-world frequency estimation applications can require dealing with extremely large vocabularies (e.g. the count of millions of elements need to be calculated). This will cause the first layer of the sketching network to have a gigantic size. For example, if a simple linear layer is used as sketching network, and input dimension (vocabulary size) is $d=10^6$, with a sketch size of $m=10^3$, the sketching layer will have $10^9$ parameters. However, note that in both training and inference, there is no need to load all parameters into memory as the sketching operation $Wx_j$ can be simply seen as a look-up table, where the index $k$ of the one-hot encoded element ($x_{jk} = 1$) indicates which row of the projection is loaded ($w_k = Wx_j$). In the scenario of multi-hot encoded inputs, the look-up operation needs to be performed as many times as non-zero elements are present in the input, followed by a simple addition of the $m$-dimensional vectors. A similar approach can be applied in the first layer of non-linear sketching networks or other multi-layer sketching networks. Such mapping is similar to the common learnt embeddings used in natural language processing applications with neural networks.

\subsubsection{Parallel sketching for large vocabularies}
Linear and non-linear sketching networks can be trained and tested without the need of loading all the network's weights providing some scalability to high dimensional inputs (large vocabularies). However, a large number of parameters still needs to be trained and stored. To further increase the scalability to large vocabularies, we propose training a SQNet with a fix input and output dimensionality and apply in parallel multiple instances of the network when larger dimensional inputs are found. Specifically, input dimensions are divided into non-overlapping windows, and the dimensions of each window are processed by a small SQNet.

\subsubsection{Ternary linear sketching for low-latency applications}
Data streaming applications using frequency estimation might require to have a very low latency. While neural network GPU-based implementations might not be able to match highly optimized low-latency frequency and membership estimation techniques, as explored in \cite{kraska2018case}, Sketch-Query Network can be used to learn a sketching function that can be later deployed into a low-latency pipeline based on Count-Sketch or Count-Min. In order to do so, we can limit the sketching function to a linear mapping with ternary values so each entry of the linear projection $W_{l,k} \in \{-1,0,1\}$ to obtain a mapping similar to a Count-Sketch, or to binary values with $W_{l,k} \in \{0,1\}$ to obtain a mapping similar to a Count-Min or Bloom Filter.

In many applications, the sketching function is performed with more frequency than the query function. For example, in data streaming scenario, every time a new sample is present, the sketch needs to be updated. However, the query function is only computed every time a user or a downstream application needs to extract information from the sketch. We can benefit of this asymmetry to use a fast Count-Sketch-like Sketching Network, while keeping a non-linear MLP as a query function to have both low-latency and an accurate reconstruction.

\subsection{Experimental Results}
\label{sec:exp_data_streaming applications}

\subsubsection{Membership estimation Zipf results}

Figure~\ref{fig:membership} shows the mean square error of a Zipf dataset with samples with $d=1000$ dimensions and sketching methods of a sketch size $m=100$, showing that SQNet surpasses BF for mildly skewed datasets, and matches its performance with almost 0 error for highly skewed datasets. Note that a Bloom Filter can perform poorly if not properly tuned (Bloom filter with W=20,D=5), obtaining higher MSE than trivially outputting always 0s (All 0s). SQNet consistently surpasses RFQNet, showing the benefits of learning the sketching function instead of using a randomized projection.

\subsubsection{Frequency and membership results with tabular datasets}

Table~\ref{tab:freq+memb} (top) shows the MSE in the datasets used for the autoencoder experiments. The table includes SQNets and RFQNets trained with Zipf dataset, as well as the networks after fine-tuning with the training sets of each respective database, showing that while networks trained with Zipf datasets already surpass traditional techniques, near zero error can be obtained if the networks are trained to match the underlying distribution of the dataset.
Similarly, Table~\ref{tab:freq+memb} (bottom) shows the results in the tabular binary datasets for membership estimation, comparing it with SQNet and RFQNet fine-tuned with each of the training sets. A similar trend as in frequency estimation is observed, showing that the error decreases significantly when the networks are fine-tuned obtaining close to 0 error. 

\begin{table*}
  \caption{Mean square error for frequency estimation (top) and membership estimation (bottom) tasks. (f) indicates that the network has been fine-tuned with each respective dataset. }
  \Description{MSE for frequency}
  \label{tab:freq+memb}
  \centering
  \begin{tabular}{l|ccccccccc|c}
    \toprule
    Model      & M & FM &  H  &  Dogs   &  KDD  & Heart  & QD  &  Ad & B  & Avg \\
    \midrule
    \midrule
    CS   & $6\mathrm{e}{-2}$ & $1\mathrm{e}{-1}$  &  $1\mathrm{e}{-1}$   &  $2\mathrm{e}{-1}$   &  $2\mathrm{e}{-1}$   &  $1\mathrm{e}{-2}$ & $4\mathrm{e}{-2}$ & $6\mathrm{e}{-3}$ & $8\mathrm{e}{-3}$  & $89\mathrm{e}{-3}$   \\
    CM   & $6\mathrm{e}{-1}$ & $6\mathrm{e}{-1}$  & $6\mathrm{e}{-1}$   &  $5\mathrm{e}{-1}$   &  $6\mathrm{e}{-1}$   &  $5\mathrm{e}{-2}$ & $6\mathrm{e}{-1}$ & $3\mathrm{e}{-2}$ & $9\mathrm{e}{-2}$ & $40\mathrm{e}{-2}$   \\
    \midrule
    RFQ   & $2\mathrm{e}{-2}$  & $7\mathrm{e}{-2}$  &  $4\mathrm{e}{-2}$   &  $8\mathrm{e}{-2}$    &  $7\mathrm{e}{-2}$   &  $1\mathrm{e}{-2}$ & $7\mathrm{e}{-3}$ & $4\mathrm{e}{-3}$ & $7\mathrm{e}{-3}$ & $36\mathrm{e}{-3}$   \\
    SQ   & $2\mathrm{e}{-2}$ & $5\mathrm{e}{-2}$  & $4\mathrm{e}{-2}$   & $7\mathrm{e}{-2}$    &  $7\mathrm{e}{-2}$   & $6\mathrm{e}{-3}$ & $7\mathrm{e}{-3}$ & $3\mathrm{e}{-3}$ & $6\mathrm{e}{-3}$ & $32\mathrm{e}{-3}$   \\
    RFQ (f)   & $5\mathrm{e}{-4}$  & $9\mathrm{e}{-4}$  &  $6\mathrm{e}{-3}$   &  $4\mathrm{e}{-2}$   &  $2\mathrm{e}{-3}$   &  $8\mathrm{e}{-5}$ & $7\mathrm{e}{-4}$ & $2\mathrm{e}{-4}$ & $3\mathrm{e}{-4}$ & $54\mathrm{e}{-4}$    \\
    SQ (f)    & $3\mathrm{e}{-4}$ & $4\mathrm{e}{-4}$  &  $2\mathrm{e}{-3}$   &  $8\mathrm{e}{-3}$   &  $9\mathrm{e}{-4}$   & $3\mathrm{e}{-5}$ & $5\mathrm{e}{-4}$ & $5\mathrm{e}{-5}$ & $1\mathrm{e}{-4}$ & $14\mathrm{e}{-4}$  \\
    \midrule
    \midrule
    BF   & $6\mathrm{e}{-1}$ & $4\mathrm{e}{-1}$  &  $6\mathrm{e}{-1}$   &  $5\mathrm{e}{-2}$   &  $2\mathrm{e}{-1}$   &  $4\mathrm{e}{-1}$ & $5\mathrm{e}{-1}$ & $4\mathrm{e}{-1}$ & $6\mathrm{e}{-1}$ & $42\mathrm{e}{-2}$    \\
    \midrule
    RFQ   & $2\mathrm{e}{-1}$ & $4\mathrm{e}{-1}$  &  $3\mathrm{e}{-1}$   &  $4\mathrm{e}{-1}$   &  $4\mathrm{e}{-1}$   &  $1\mathrm{e}{-1}$ & $3\mathrm{e}{-1}$ & $1\mathrm{e}{-1}$ & $1\mathrm{e}{-1}$ & $26\mathrm{e}{-2}$     \\
    SQ   & $2\mathrm{e}{-1}$ & $3\mathrm{e}{-1}$  &  $2\mathrm{e}{-1}$   &  $5\mathrm{e}{-2}$   &  $2\mathrm{e}{-1}$   &  $4\mathrm{e}{-2}$ & $2\mathrm{e}{-1}$ & $4\mathrm{e}{-2}$ & $7\mathrm{e}{-2}$ & $15\mathrm{e}{-2}$    \\
    RFQ (f)   & $7\mathrm{e}{-2}$ & $6\mathrm{e}{-2}$  &  $1\mathrm{e}{-1}$   &  $1\mathrm{e}{-1}$   &  $2\mathrm{e}{-1}$   &  $2\mathrm{e}{-2}$ & $1\mathrm{e}{-1}$ & $2\mathrm{e}{-2}$ & $4\mathrm{e}{-2}$ & $86\mathrm{e}{-3}$     \\
    SQ (f)    & $1\mathrm{e}{-2}$ & $2\mathrm{e}{-2}$  &  $1\mathrm{e}{-1}$   &  $1\mathrm{e}{-2}$   &  $2\mathrm{e}{-1}$   &  $5\mathrm{e}{-3}$ & $1\mathrm{e}{-1}$ & $6\mathrm{e}{-3}$ & $2\mathrm{e}{-2}$ & $64\mathrm{e}{-3}$   \\
    \bottomrule
  \end{tabular}
\end{table*}

\subsubsection{Performance evaluation for frequency estimation}

We evaluate the computational time between different SQNet configurations and CS, CM, and RFQNet.  Every method is trained and evaluated in a V100 GPU. Table~\ref{tab:mem_upddated} shows the runtimes for training and testing of each of the methods. CS and CM do not require any training and provide the fastest inference time, while providing a higher reconstruction error. RFQNet and SQNet using Linear and Ternary layers as Sketching Networks, provide faster inference times than using a non-linear SQNet.

Note that real-world applications of frequency and membership estimation methods might require very low latencies that might be challenging to obtain by using neural networks implemented in GPU (See discussion in \cite{kraska2018case}). 

\begin{table*}
   \caption{Train and test time of each method in seconds.}
   \Description{time}
   \label{tab:mem_upddated}
   \centering
   \begin{tabular}{l|ccccccccc|c}
     \toprule
     Model      & CM & CS &  SQ(Non Linear) &  SQ(Linear) &  SQ(Ternary) &  SQ(Parallel)  &  RFQ \\
     \midrule
     \midrule
     Train   & - & -  &  $1\mathrm{e}{3}$  & $4\mathrm{e}{2}$ & $2\mathrm{e}{3}$  & $2\mathrm{e}{3}$ &  $5\mathrm{e}{2}$  \\
     Test   & $5\mathrm{e}{-4}$ & $6\mathrm{e}{-4}$   &  $1\mathrm{e}{-1}$ & $3\mathrm{e}{-3}$ & $3\mathrm{e}{-3}$  & $6\mathrm{e}{-2}$ & $3\mathrm{e}{-3}$  \\
     \bottomrule
   \end{tabular}
\end{table*}

\subsubsection{Hyperparameter search for Count-Min and Count-Sketch}

We explore the optimal hyperparameters for Count-Min and Count-Sketch. Both methods have two main parameters: $N_w$, which is the output dimension of the random linear projections, and $N_d$, which is the number of projections. Both methods generate sketches of size $m=N_wN_d$, therefore, for a fair comparison, we fix the sketch size for all methods (e.g. $m=100$) and look for the combination of parameters $N_w$ and $N_d$ (e.g. such as $N_wN_d=100$) that provide the lowest mean square error.

To perform the hyperparameter search for CM and CS, we generate Zipf datasets with multiple different combinations of $\alpha$ and $\beta$. We explore $\alpha$ ranging from 0 to 2 and $\beta$ ranging from 0 to 1. For each value of $\alpha$ and $\beta$, we try all possible values of $N_d$ and $N_w$ (making sure that $N_wN_d=m$). We evaluate each setting 5 times and compute the average mean square error and its standard deviation with a dataset that for each $\alpha$ and $\beta$ includes 500 batches with each including a set of 100 samples with 1000 dimensions each. (The input of the sketching method has dimensionality $500 \times 100 \times 1000$, the input of the decoding method (i.e. sketches) has dimensionality $500 \times m$, and the predicted query $500 \times 1000$.) We search the hyperparameters with a sketch size of $m=100$ and $m=10$. 

\begin{figure}[ht]
  \centering
  \includegraphics[width=1\linewidth]{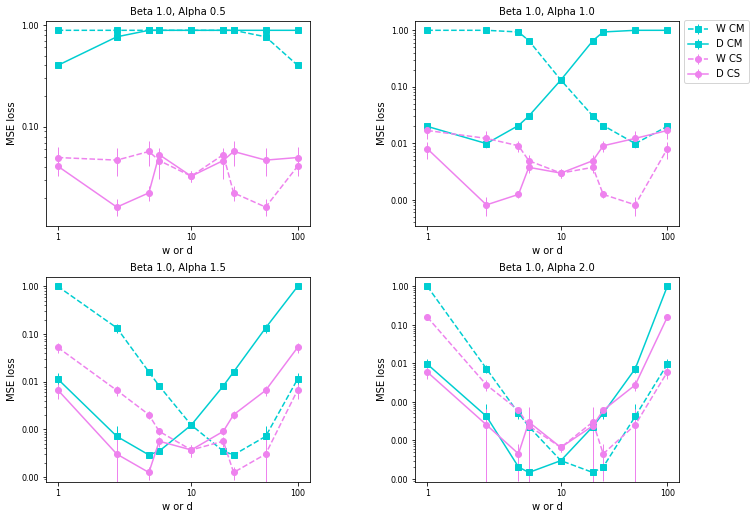}
  \caption{MSE mean and standard deviation with ten executions of CM and CS using different d and w, with a sketch size of 100. Using Zipf-distributed dataset for different $\alpha$ values and $\beta = 1.0$.}
  \Description{MSE Zipf}
  \label{fig:frequency_2}
\end{figure}

Figure~\ref{fig:frequency_2} shows the mean and standard deviation of the Mean Square Error (MSE) for Count-Min (CM) and Count-Sketch (CS) using different values of $N_d$ and $N_w$. Each plot corresponds to a particular $\alpha$ and $\beta$, and both x and y axes are in a log scale. Note that the optimal configurations of CS outperform the optimal configurations of CM, except in extremely skewed settings ($\alpha = 2$). For low values of $\alpha$, the collision between elements within the sketch is large, leading to high errors in CM, regardless of the parameters selected. For skewed datasets ($\alpha > 1$) the selection of good parameters becomes more critical. For example, if poor parameters are selected (e.g. $N_w=1, N_d=100$) both CM and CS perform more than four orders of magnitude worse than if suitable parameters are selected (e.g. $N_w=25, N_d=4$). Note that small values of $N_w$ will lead to high number of collisions (i.e. many input elements getting projected into the same sketch dimension), and increasing $N_d$ won't help much (e.g. $N_w=1, N_d=100$ always performs poorly). However, if $N_w$ has a medium or large dimension, collision numbers are decreased and increasing $N_d$ (increasing the number of projections) provides a significant improvement.

Figure~\ref{fig:frequency_3} shows the optimal value of $N_w$ for both CS, CM and Bloom Filter (discussed in the next section) for different values of $\alpha$ and $\beta$. Because $N_wN_d=100$, the parameter of $N_d$ can be easily inferred with $N_d=100/N_w$. Note that for small $\alpha$, CM works optimally with $N_w=100, N_d=1$, as low dimensional projections lead to a high number collisions and bad predictions. When $\alpha>1$, both CM and CS perform better with a smaller size of $N_w$ and larger $N_d$. Figure~\ref{fig:frequency_4} provides the MSE of the optimal configurations of both CM and CS for different values of $\alpha$ and $\beta$. Following the same trend as in Figures \ref{fig:frequency}, \ref{fig:frequency_2}, \ref{fig:frequency_3}, we can observe that CM performs poorly for $\alpha < 1$, while CS provides more robustness to non-skewed distributions, with an error decreasing for more sparse data (lower error for smaller values of $\beta$).

\subsubsection{Hyperparameter search for Bloom Filter}

\begin{figure}[htpb]
  \centering
  \includegraphics[width=1\linewidth]{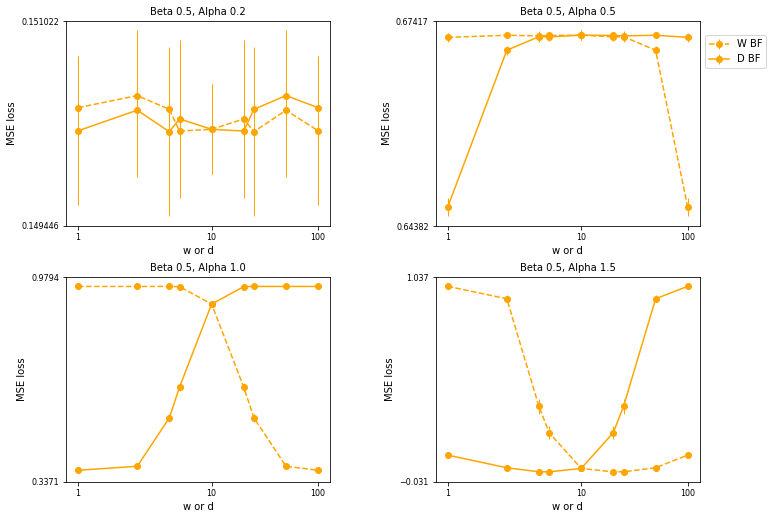}
  \caption{MSE mean and standard deviation with 10 executions of Bloom Filters using different d and w, with a sketch size of 100. Using Zipf-distributed dataset for different $\alpha$ values and $\beta = 0.5$.}
  \Description{MSE with 10 execs}
  \label{fig:membership_1}
\end{figure}

We apply a similar hyperparameter search for Bloom Filters (BF) as performed for CM and CS. Zipf-datasets with different values of $\alpha$ and $\beta$ are used to evaluate bloom filters. Here we perform ten runs where for each $\alpha$ $\beta$ pair a total of 500 batches with each including a set with ten 1000-dimensional binary inputs are used to evaluate bloom filter.
Figure~\ref{fig:membership_1} shows the mean and standard deviation of the Mean Square Error (MSE) of BF using different values of $N_d$ and $N_w$. Each plot corresponds to a particular $\alpha$ and $\beta$, and we can see how the MSE varies between each run. A similar behavior as CM is observed in BF, with the method performing poorly for small $\alpha$ and obtaining a good performance for large $\alpha$, if hyperparameters are selected properly.

\begin{figure}[ht]
  \centering
  \includegraphics[width=1\linewidth]{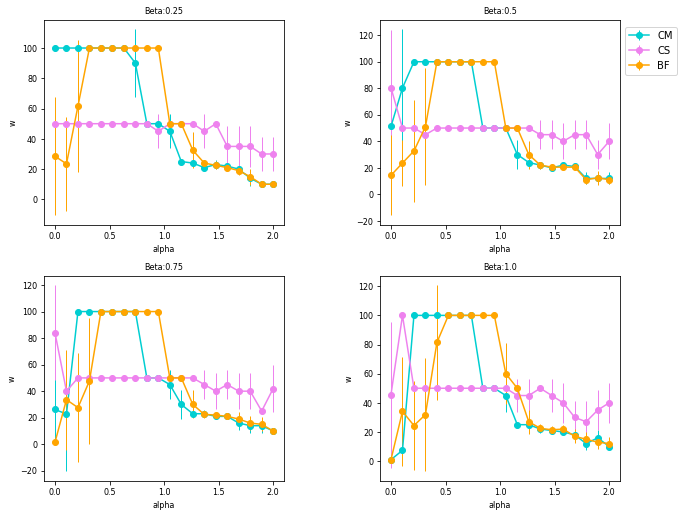}
  \caption{Mean and standard deviation of the optimal $w$ from ten runs using Count-Min, Count-Sketch and Bloom Filters (sketch size of 100) for the Zipf-distributed dataset with different $\alpha$ and $\beta$ values.}
  \Description{Optimal w}
  \label{fig:frequency_3}
\end{figure}

\begin{figure}[ht]
  \centering
  \includegraphics[width=1\linewidth]{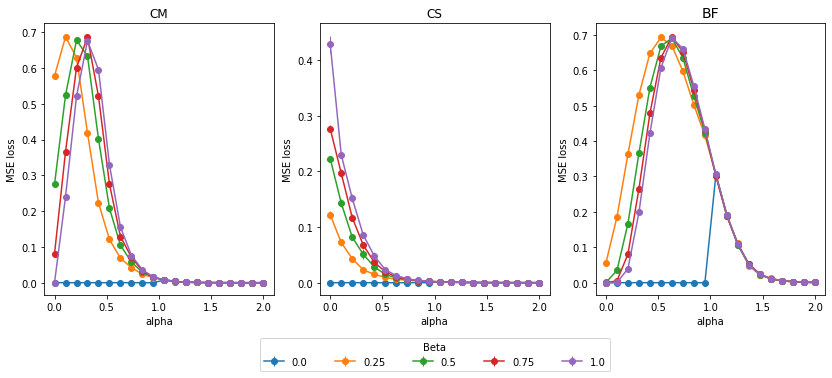}
  \caption{MSE mean and standard deviation of 10 executions of Count Min Sketch, Count Sketch, and Bloom Filters with the optimum d and w, with a sketch size of one hundred. Using Zipf-distributed dataset for different $\alpha$ and $\beta$.}
  \Description{MSE using count min sketch}
  \label{fig:frequency_4}
\end{figure}

Figure~\ref{fig:frequency_3} (orange line), the optimal value of $w$ for BF is shown for different values of $\alpha$ and $\beta$ showing a similar behaviour with CM. The similarity with CM can be further observed in Figure~\ref{fig:frequency_4} (right), which provides the MSE of BF for different $\alpha$ and $\beta$. The optimal values of $N_d$ and $N_w$ for each $\alpha$ and $\beta$ are used to compute the errors of BF shown in Figure~\ref{fig:membership}.

\subsubsection{Hyperparameter search for SQNet and RFQNet for frequency estimation}

Both SQNet and RFQNet are trained using random batches following the Zipf distribution. We train the networks with a batch size of 500, where each batch includes a set of multiple 1000-dimensional samples. The size of the set is randomly selected at every iteration with values between one and twenty elements. Furthermore, the $\alpha$ and $\beta$ values are randomly chosen with $\alpha$ taking values from zero to two, and $\beta$ from zero to one.

\begin{figure}[h]
  \centering
  \includegraphics[width=1\linewidth]{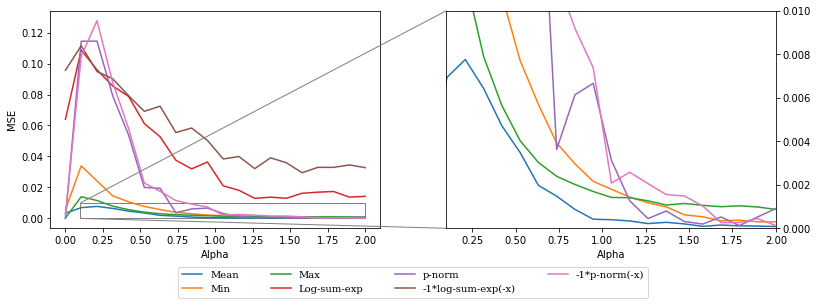}
  \caption{Mean square error (MSE) for frequency estimation with SQNet in a Zipf-distributed dataset for different $\alpha$ values and pooling layers with $\beta = 1.0$. Right plot shows a zoomed–in view of the MSE range [0,0.01]}
  \Description{MSE for frequency in a Zipf-distributed dataset}
  \label{fig:frequency_5}
\end{figure}

Note that in SQNet, both the Sketch and the Query network are trained, while in RFQNet, only the Query Network is trained and the Sketch Network is replaced by Random Features. For SQNet we explore two different architectures for the Sketch Network, which are described in Figure~\ref{fig:net_arch}. For frequency estimation task, the architecture that provided better results in most settings is the ``ResNet'' style architecture (Network (a) in Figure~\ref{fig:net_arch}).
We explore different pooling layers, learning rates, hidden sizes, number of hidden layers and pooling layer parameters such as $p$ values between one and ten for the p-norm pooling layers. Figure~\ref{fig:frequency_5} shows the MSE obtained with networks with different pooling layers. For each pooling layer, we report the results from the best performing hyperparameters in a validation Zipf dataset generated in a similar way as the training set. The Mean pooling layer (blue line) outperforms all other pooling operations, with log-sum-exp and p-norm based pooling layers performing poorly. Because the pair of Sketch and Query Network is approximating an average function (i.e. averaging a set to obtain its normalized frequency), it is expected that the mean pooling layer performs well.

\begin{figure}[h]
  \centering
  \includegraphics[width=1\linewidth]{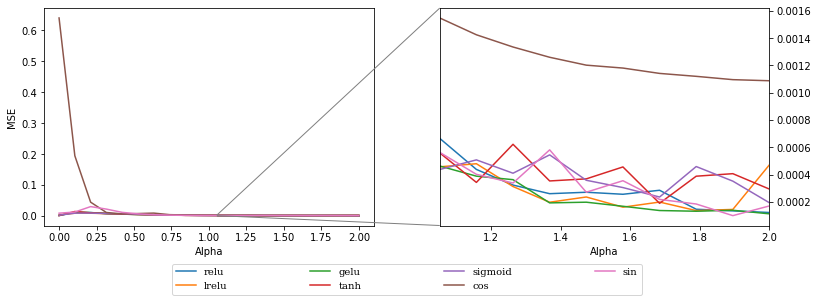}
  \caption{Mean square error (MSE) for frequency estimation with RFQNet in a Zipf-distributed dataset for different $\alpha$ values and activations layers with $\beta = 1.0$. Right plot shows a zoomed–in view of the alpha range [1.0,2.0]}
  \Description{MSE for frequency with RFQNet in a Zipf-distributed dataset}
  \label{fig:frequency_6}
\end{figure}

For RFQNet, we fix the Mean as a pooling layer (as commonly done in CL appliations) and we explore different type of random features (with different activations) and variances. The range of values of the variances goes from 0.005 to 1 and the activations are the ReLU, GELU, LeakyReLU, Fourier Features (Cosine and Sine), Cosine, Sine, Tanh, and Sigmoid. Note that each activation will lead to random features that approximate a different kernel. A comparison between the activations using the best performing variances can be seen in Figure~\ref{fig:frequency_6}. The best performing random features are the GELU features, with Cosine and Sigmoid features performing the worst.

\begin{figure}[h]
  \centering
  \includegraphics[width=1\linewidth]{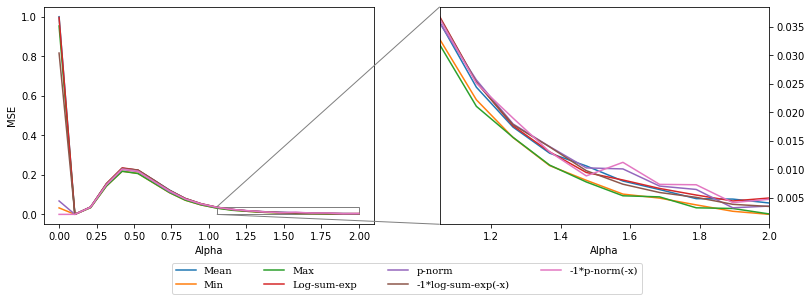}
  \Description{MSE for membership in a Zipf-distributed dataset}
  \caption{Mean square error (MSE) for membership estimation with SQNet in a Zipf-distributed dataset for different $\alpha$ values and pooling layers with $\beta = 0.5$. Right plot shows a zoomed–in view of the MSE range [0,0.01]}
  \label{fig:membership_4}
\end{figure}

\begin{figure}[h]
  \centering
  \includegraphics[width=1\linewidth]{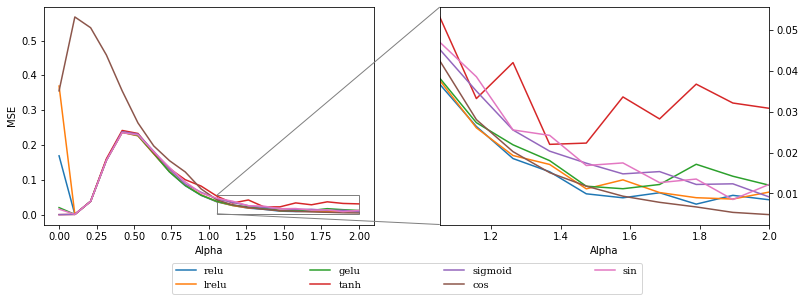}
  \caption{Mean square error (MSE) for membership estimation with RFQNet in a Zipf-distributed dataset for different $\alpha$ values and activations layers with $\beta = 0.5$. Right plot shows a zoomed–in view of the alpha range [1.0,2.0]}
  \Description{MSE for membership with RFQNet in a Zipf-distributed dataset}
  \label{fig:membership_5}
\end{figure}

\subsubsection{Hyperparameter search for SQNet and RFQNet for membership estimation}

For membership estimation, we adopt the same training procedure as in frequency estimation with minor changes. For simplicity, we fix during training and testing both the number of samples within each set to 10, and the $\beta$ value to 0.5. As in frequency estimation, the batch size is 500 and the value of $\alpha$ is randomly selected between zero and two, with some batches randomly limiting its range from 0.75 to 1.5 as those were the regions where the networks were under-performing.

As in frequency estimation, we explore two different architectures, and different pooling layers for the Sketch Network. In this framework, some pooling layers work better with the architecture (b) described in Figure~\ref{fig:net_arch}, in particular the Mean, the Minimum, the Maximum, and the Log-Sum-Exp. The other pooling layers work better with the architecture (a). In addition, we explore different learning rates, number of hidden layers, and hidden sizes. In Figure~\ref{fig:membership_4} there is a comparison between the different pooling layers using the best combination of hyperparameters with a $\beta = 0.5$. The best pooling layer is the Minimum with the architecture (b), obtaining a performance very similar to the maximum operation. Note that in this application the mean pooling layer is one of the pooling layers providing the highest MSE. 
In the RFQnet, we explore several activation layers and variances. The activations are the same as in the frequency estimation. Figure~\ref{fig:membership_5} shows a comparison between activation layers using the best hyperparameters and with a $\beta = 0.5$. The best activation layer is GELU with a variance of 0.6.

\begin{table*}[!ht]
\caption{Meta-training datasets for PCA experiments. Subscripts $i..j$ and $(\cdot)$ denote the interval of indices and the total number of datasets of the same family used, respectively. Train and test subsets are combined into a single set.}
\Description{List of meta-training datasets}
\label{resultstable1}
\begin{center}
\begin{small}
    \begin{tabular}{lccc}
    \toprule
        \text{Dataset name} & \text{\# Classes} & \text{\# Features} & \text{\# Samples}\\
        \midrule
        Fashion-MNIST & 10 & 784 & 70000 \\ 
        EMNIST-digits & 10 & 784 & 280000 \\ 
        letters$_1$ & 10 & 784 & 56000 \\ 
        quickdraw & 10 & 784 & 100000 \\ 
        hapmap$_{1..30}$ (30) & 10 & 784 & 2214 \\ 
        dogstrain$_{1..14}$ (14) & 10 & 784 & 1830 \\ 
        Bioresponse & 2 & 1776 & 3751 \\ 
        splice & 3 & 60 & 3190 \\ 
        qsar-biodeg & 2 & 41 & 1055 \\ 
        MiceProtein & 8 & 77 & 1080 \\ 
        kr-vs-kp & 2 & 36 & 3196 \\ 
        churn & 2 & 20 & 5000 \\ 
        CIFAR-10 & 10 & 3072 & 60000 \\ 
        kc1 & 2 & 21 & 2109 \\ 
        jungle-chess & 3 & 6 & 44819 \\ 
        adult & 2 & 14 & 48842 \\ 
        nomao & 2 & 118 & 33465 \\ 
        first-order-theorem-proving & 6 & 51 & 6118 \\ 
        Devnagari-Script & 46 & 1024 & 92000 \\ 
        mfeat-morphological & 10 & 6 & 2000 \\ 
        pc3 & 2 & 37 & 1563 \\ 
        vehicle & 4 & 18 & 846 \\ 
        vowel & 11 & 12 & 990 \\ 
        letter & 26 & 16 & 20000 \\ 
        ozone-level-8hr & 2 & 72 & 2534 \\ 
        spambase & 2 & 57 & 4601 \\ 
        PhishingWebsites & 2 & 30 & 11055 \\ 
        climate-model-simulation-crashes & 2 & 18 & 540 \\ 
        banknote-authentication & 2 & 4 & 1372 \\ 
        credit-approval & 2 & 15 & 690 \\ 
        bank-marketing & 2 & 16 & 45211 \\ 
        cmc & 3 & 9 & 1473 \\ 
        phoneme & 2 & 5 & 5404 \\  
        wdbc & 2 & 30 & 569 \\ 
        eucalyptus & 5 & 19 & 736 \\ 
        dresses-sales & 2 & 12 & 500 \\ 
        breast-w & 2 & 9 & 699 \\ 
    \bottomrule
    \end{tabular}
\end{small}
\end{center}
\end{table*}

\begin{table*}[!ht]
\caption{Meta-testing datasets for PCA experiments. Subscripts $i..j$ and $(\cdot)$ denote the interval of indices and the total number of datasets of the same family used, respectively. Train and test subsets are combined into a single set.}
\Description{List of meta-testing datasets}
\label{resultstable2}
\begin{center}
\begin{small}
    \begin{tabular}{lccc}
    
    \toprule
        \text{Dataset name} & \text{\# Classes} & \text{\# Features} & \text{\# Samples}\\ 
        \midrule
        MNIST & 10 & 784 & 70000 \\ 
        letters$_2$ & 10 & 784 & 56000 \\ 
        KMNIST & 10 & 784 & 70000 \\ 
        ethiopic & 10 & 784 & 70000 \\ 
        osmanya & 10 & 784 & 70000 \\ 
        vai & 10 & 784 & 70000 \\ 
        hapmap$_{31..50}$ (20) & 10 & 784 & 2214 \\ 
        dogstrain$_{15..30}$ (16) & 10 & 784 & 1830 \\ 
        sick & 2 & 29 & 3722 \\ 
        pendigits & 10 & 16 & 10992 \\ 
        isolet & 26 & 617 & 7797 \\ 
        connect-4 & 3 & 42 & 67557 \\ 
        analcatdata-authorship & 4 & 70 & 841 \\ 
        analcatdata-dmft & 6 & 4 & 797 \\ 
        cylinder-bands & 2 & 37 & 540 \\ 
        optdigits & 10 & 64 & 5620 \\ 
        wall-robot-navigation & 4 & 24 & 5456 \\ 
        mfeat-karhunen & 10 & 64 & 2000 \\ 
        mfeat-factors & 10 & 216 & 2000 \\ 
        texture & 11 & 40 & 5500 \\ 
        pc4 & 2 & 37 & 1458 \\ 
        satimage & 6 & 36 & 6430 \\ 
        Internet-Advertisements & 2 & 1558 & 3279 \\ 
        har & 6 & 561 & 10299 \\ 
        diabetes & 2 & 8 & 768 \\ 
        segment & 7 & 16 & 2310 \\ 
        car & 4 & 6 & 1728 \\ 
        credit-g & 2 & 20 & 1000 \\ 
        wilt & 2 & 5 & 4839 \\ 
        jm1 & 2 & 21 & 10885 \\ 
        kc2 & 2 & 21 & 522 \\ 
        numerai28.6 & 2 & 21 & 96320 \\ 
        dna & 3 & 180 & 3186 \\ 
        mfeat-zernike & 10 & 47 & 2000 \\ 
        madelon & 2 & 500 & 2600 \\ 
        blood-transfusion-service-center & 2 & 4 & 748 \\ 
        balance-scale & 3 & 4 & 625 \\ 
        electricity & 2 & 8 & 45312 \\ 
        pc1 & 2 & 21 & 1109 \\ 
        tic-tac-toe & 2 & 9 & 958 \\ 
        semeion & 10 & 256 & 1593 \\ 
        GesturePhaseSegmentationProcessed & 5 & 32 & 9873 \\
        cnae-9 & 9 & 856 & 1080 \\ 
        ilpd & 2 & 10 & 583 \\ 
        mfeat-fourier & 10 & 76 & 2000 \\ 
        steel-plates-fault & 7 & 27 & 1941 \\ 
        mfeat-pixel & 10 & 240 & 2000 \\ 
    \bottomrule
    \end{tabular}
\end{small}
\end{center}
\end{table*}

\begin{table*}[!ht]
\caption{Reconstruction error for each test dataset with Sketch-Query Network for different sketch sizes (Part 1).}
\Description{Results on each test dataset with different sizes part 1}
\label{resultstable3_part1}
\begin{center}
\setlength\tabcolsep{4pt}
\begin{small}
    \begin{tabular}{lcccccccccc}
    \toprule
        Dataset & 2 & 10 & 20 & 100 & 200 & 500 & 1000 & 2000 & 15000 & 19306 \\ 
        \midrule
        letters$_2$ & 81.82 & 82.001 & 81.909 & 81.961 & 81.795 & 80.15 & 77.018 & 74.933 & 63.56 & 56.613 \\ \hline
        KMNIST & 92.953 & 92.825 & 92.949 & 92.613 & 92.733 & 90.395 & 89.25 & 86.761 & 76.709 & 71.377 \\ \hline
        ethiopic & 62.125 & 62.56 & 61.972 & 63.56 & 63.782 & 61.46 & 60.524 & 60.097 & 53.135 & 50.936 \\ \hline
        osmanya & 67.142 & 67.103 & 67.847 & 66.765 & 68.32 & 65.129 & 64.406 & 61.703 & 53.139 & 49.39 \\ \hline
        vai & 71.465 & 70.484 & 71.418 & 70.788 & 70.784 & 69.37 & 67.196 & 65.798 & 56.91 & 52.911 \\ \hline
        hapmap31 & 84.963 & 83.258 & 85.661 & 84.84 & 82.812 & 79.871 & 77.452 & 74.353 & 69.641 & 58.107 \\ \hline
        hapmap32 & 85.004 & 85.259 & 85.197 & 85.0 & 81.605 & 78.759 & 77.963 & 75.484 & 70.316 & 59.166 \\ \hline
        hapmap33 & 84.74 & 83.424 & 84.928 & 82.306 & 81.915 & 78.653 & 76.967 & 74.871 & 70.09 & 59.459 \\ \hline
        hapmap34 & 86.086 & 85.297 & 85.727 & 84.778 & 81.754 & 79.68 & 77.322 & 75.226 & 69.992 & 59.186 \\ \hline
        hapmap35 & 86.59 & 84.957 & 85.67 & 85.213 & 82.184 & 79.245 & 77.785 & 75.702 & 69.984 & 59.432 \\ \hline
        hapmap36 & 84.609 & 84.302 & 83.613 & 84.08 & 81.782 & 78.17 & 77.158 & 74.706 & 69.643 & 58.568 \\ \hline
        hapmap37 & 86.018 & 84.975 & 84.942 & 85.341 & 81.693 & 78.686 & 76.796 & 74.838 & 70.378 & 58.681 \\ \hline
        hapmap38 & 86.17 & 85.325 & 84.894 & 82.77 & 82.162 & 79.058 & 77.356 & 74.471 & 70.082 & 58.853 \\ \hline
        hapmap39 & 85.943 & 84.46 & 85.651 & 84.372 & 82.407 & 79.161 & 76.158 & 75.443 & 70.575 & 58.958 \\ \hline
        hapmap40 & 84.845 & 85.808 & 84.388 & 84.578 & 81.68 & 79.731 & 76.946 & 75.228 & 70.439 & 59.596 \\ \hline
        hapmap41 & 86.194 & 85.408 & 83.298 & 83.226 & 82.424 & 78.151 & 76.693 & 75.694 & 70.214 & 58.638 \\ \hline
        hapmap42 & 85.682 & 85.173 & 86.591 & 83.314 & 82.046 & 78.07 & 76.888 & 74.651 & 69.565 & 59.157 \\ \hline
        hapmap43 & 86.191 & 85.045 & 85.127 & 84.689 & 82.33 & 79.121 & 77.762 & 74.811 & 70.576 & 58.732 \\ \hline
        hapmap44 & 86.131 & 85.377 & 84.789 & 84.254 & 82.74 & 79.496 & 76.761 & 75.59 & 69.861 & 59.417 \\ \hline
        hapmap45 & 86.38 & 85.804 & 84.862 & 84.708 & 81.305 & 79.433 & 77.495 & 75.16 & 69.843 & 58.77 \\ \hline
        hapmap46 & 86.567 & 83.789 & 85.116 & 83.282 & 81.714 & 78.56 & 77.322 & 74.636 & 69.884 & 58.925 \\ \hline
        hapmap47 & 85.907 & 84.696 & 84.836 & 84.062 & 82.874 & 78.945 & 77.503 & 74.984 & 70.566 & 58.801 \\ \hline
        hapmap48 & 84.752 & 85.474 & 85.039 & 83.599 & 80.604 & 78.153 & 77.142 & 74.513 & 70.331 & 59.715 \\ \hline
        hapmap49 & 84.694 & 84.939 & 85.018 & 83.755 & 82.746 & 78.882 & 76.61 & 75.421 & 69.215 & 59.373 \\ \hline
        hapmap50 & 85.774 & 85.055 & 83.829 & 84.041 & 80.899 & 77.809 & 76.962 & 74.691 & 68.392 & 58.266 \\ \hline
        dogstrain15 & 87.953 & 88.264 & 87.811 & 87.254 & 86.708 & 84.983 & 83.098 & 80.147 & 67.796 & 59.919 \\ \hline
        dogstrain16 & 88.056 & 88.061 & 87.644 & 87.738 & 86.365 & 84.706 & 83.096 & 79.901 & 67.901 & 59.756 \\ \hline
        dogstrain17 & 88.48 & 88.9 & 88.247 & 87.494 & 87.147 & 84.701 & 83.427 & 79.571 & 68.028 & 59.568 \\ \hline
        dogstrain18 & 88.653 & 88.588 & 88.278 & 87.819 & 86.572 & 84.673 & 83.073 & 79.752 & 68.039 & 59.735 \\ \hline
        dogstrain19 & 88.569 & 88.124 & 88.724 & 87.58 & 87.05 & 84.718 & 83.311 & 80.65 & 68.061 & 59.615 \\ \hline
        dogstrain20 & 88.237 & 88.383 & 87.692 & 88.149 & 87.113 & 84.813 & 83.466 & 79.933 & 67.729 & 59.69 \\ \hline
        dogstrain21 & 88.133 & 88.459 & 88.019 & 87.474 & 86.965 & 84.993 & 82.947 & 80.417 & 68.05 & 60.079 \\ \hline
        dogstrain22 & 88.554 & 88.26 & 87.981 & 87.973 & 86.865 & 84.507 & 82.507 & 80.046 & 67.981 & 59.224 \\ \hline
        dogstrain23 & 88.43 & 87.845 & 88.838 & 87.374 & 86.856 & 84.887 & 83.282 & 80.361 & 68.153 & 59.304 \\ \hline
        dogstrain24 & 88.803 & 88.726 & 88.265 & 87.528 & 87.088 & 84.966 & 82.933 & 79.298 & 67.631 & 60.133 \\ \hline
        dogstrain25 & 88.367 & 88.669 & 88.25 & 87.548 & 86.394 & 84.721 & 83.194 & 79.869 & 68.095 & 59.29 \\ \hline
        dogstrain26 & 88.423 & 88.402 & 88.23 & 87.682 & 87.172 & 84.995 & 83.463 & 80.469 & 67.929 & 59.641 \\ \hline
        dogstrain27 & 88.599 & 88.944 & 88.236 & 87.5 & 86.674 & 84.405 & 83.079 & 80.245 & 67.856 & 59.742 \\ \hline
        dogstrain28 & 88.498 & 87.82 & 88.147 & 87.658 & 86.246 & 83.883 & 82.9 & 79.935 & 67.829 & 59.991 \\ \hline
        dogstrain29 & 89.103 & 88.436 & 88.993 & 87.833 & 86.869 & 84.866 & 83.547 & 80.058 & 67.942 & 60.336 \\ \hline
        dogstrain30 & 88.077 & 88.308 & 87.959 & 87.686 & 87.178 & 84.98 & 84.385 & 80.064 & 68.189 & 59.247 \\
    \bottomrule
    \end{tabular}
\end{small}
\end{center}
\end{table*}

\begin{table*}[!ht]
\caption{Reconstruction error for each test dataset with Sketch-Query Network for different sketch sizes (Part 2).}
\Description{Results on each test dataset with different sizes part 2}
\label{resultstable3_part2}
\begin{center}
\setlength\tabcolsep{4pt}
\begin{small}
    \begin{tabular}{lcccccccccc}
    \toprule
        Dataset & 2 & 10 & 20 & 100 & 200 & 500 & 1000 & 2000 & 15000 & 19306 \\ 
        \midrule
        sick & 1.779 & 1.713 & 1.797 & 1.705 & 1.667 & 1.53 & 1.459 & 1.408 & 1.2 & 1.067 \\ \hline
        pendigits & 0.614 & 0.63 & 0.693 & 0.639 & 0.609 & 0.564 & 0.601 & 0.495 & 0.244 & 0.187 \\ \hline
        isolet & 95.361 & 92.762 & 92.047 & 91.137 & 86.432 & 78.733 & 73.121 & 67.621 & 48.206 & 28.783 \\ \hline
        connect-4 & 4.309 & 3.642 & 3.525 & 3.714 & 3.616 & 3.398 & 3.456 & 3.267 & 2.715 & 2.5 \\ \hline
        analcatdata-authorship9 & 12.215 & 12.173 & 12.109 & 11.084 & 10.992 & 10.204 & 9.788 & 8.919 & 7.011 & 6.009 \\ \hline
        analcatdata-dmft & 0.024 & 0.022 & 0.027 & 0.018 & 0.017 & 0.017 & 0.018 & 0.018 & 0.016 & 0.016 \\ \hline
        cylinder-bands & 2.739 & 2.802 & 2.851 & 2.589 & 2.457 & 2.667 & 2.557 & 2.323 & 1.665 & 1.506 \\ \hline
        optdigits & 9.668 & 8.98 & 8.452 & 8.114 & 8.206 & 7.828 & 7.407 & 6.563 & 4.419 & 3.705 \\ \hline
        wall-robot-navigation & 1.449 & 1.268 & 1.156 & 1.387 & 1.397 & 1.318 & 1.242 & 1.127 & 0.854 & 0.751 \\ \hline
        mfeat-karhunen & 10.049 & 9.88 & 10.102 & 9.8 & 9.547 & 8.884 & 8.011 & 7.37 & 5.745 & 5.206 \\ \hline
        mfeat-factors & 94.695 & 93.217 & 89.874 & 84.274 & 76.066 & 63.446 & 55.125 & 43.893 & 24.106 & 6.802 \\ \hline
        texture & 1.151 & 0.943 & 0.971 & 0.993 & 0.945 & 0.886 & 0.826 & 0.79 & 0.536 & 0.122 \\ \hline
        pc4 & 3.402 & 3.854 & 2.982 & 2.449 & 2.418 & 1.042 & 1.077 & 1.025 & 0.861 & 0.613 \\ \hline
        satimage & 4.716 & 1.747 & 1.947 & 1.707 & 1.676 & 1.609 & 1.531 & 1.293 & 0.47 & 0.187 \\ \hline
        Internet-Advertisements & 96.708 & 95.859 & 96.139 & 94.367 & 93.909 & 89.923 & 86.81 & 81.314 & 57.009 & 43.163 \\ \hline
        har & 69.804 & 69.891 & 69.316 & 61.779 & 53.622 & 51.775 & 46.926 & 45.962 & 38.472 & 23.083 \\ \hline
        diabetes & 0.115 & 0.109 & 0.105 & 0.093 & 0.1 & 0.106 & 0.092 & 0.085 & 0.083 & 0.078 \\ \hline
        segment & 0.619 & 0.625 & 0.587 & 0.496 & 0.481 & 0.517 & 0.466 & 0.403 & 0.199 & 0.124 \\ \hline
        car & 0.064 & 0.064 & 0.064 & 0.064 & 0.064 & 0.064 & 0.064 & 0.064 & 0.064 & 0.064 \\ \hline
        credit-g & 0.918 & 0.918 & 0.858 & 0.819 & 0.821 & 0.824 & 0.83 & 0.797 & 0.705 & 0.677 \\ \hline
        wilt & 0.044 & 0.032 & 0.054 & 0.033 & 0.039 & 0.028 & 0.025 & 0.024 & 0.023 & 0.021 \\ \hline
        jm1 & 0.465 & 0.48 & 0.437 & 1.437 & 1.26 & 1.229 & 0.217 & 0.208 & 0.181 & 0.143 \\ \hline
        kc2 & 0.32 & 1.92 & 0.613 & 0.507 & 0.543 & 0.448 & 0.077 & 0.069 & 0.066 & 0.047 \\ \hline
        numerai28.6 & 1.0 & 0.93 & 0.992 & 0.906 & 0.859 & 0.829 & 0.803 & 0.656 & 0.423 & 0.322 \\ \hline
        dna & 82.035 & 76.077 & 75.716 & 75.176 & 75.053 & 73.455 & 72.484 & 70.341 & 58.116 & 51.701 \\ \hline
        mfeat-zernike & 5.02 & 4.61 & 4.549 & 4.135 & 4.082 & 3.972 & 3.729 & 3.201 & 1.487 & 0.928 \\ \hline
        madelon & 97.05 & 96.761 & 96.771 & 96.34 & 96.281 & 94.609 & 93.526 & 91.604 & 84.172 & 78.698 \\ \hline
        blood-transfusion-service-center & 0.022 & 0.02 & 0.017 & 0.024 & 0.023 & 0.015 & 0.008 & 0.015 & 0.008 & 0.006 \\ \hline
        balance-scale & 0.023 & 0.023 & 0.023 & 0.023 & 0.023 & 0.023 & 0.023 & 0.023 & 0.023 & 0.023 \\ \hline
        electricity & 0.128 & 0.137 & 0.134 & 0.136 & 0.128 & 0.094 & 0.089 & 0.083 & 0.077 & 0.071 \\ \hline
        pc1 & 0.663 & 1.947 & 0.424 & 1.68 & 1.521 & 0.733 & 0.181 & 0.165 & 0.153 & 0.118 \\ \hline
        tic-tac-toe & 0.174 & 0.168 & 0.156 & 0.156 & 0.153 & 0.148 & 0.147 & 0.142 & 0.139 & 0.137 \\ \hline
        semeion & 94.128 & 93.318 & 95.104 & 94.095 & 93.3 & 82.355 & 79.416 & 70.328 & 48.351 & 31.014 \\ \hline
        GesturePhaseSegmentationProcessed & 2.386 & 2.219 & 2.32 & 2.255 & 2.232 & 2.195 & 2.131 & 1.984 & 1.228 & 1.085 \\ \hline
        cnae-9 & 96.042 & 96.437 & 96.238 & 95.997 & 95.31 & 91.713 & 89.857 & 85.756 & 69.218 & 57.001 \\ \hline
        MNIST & 79.429 & 79.083 & 79.278 & 79.073 & 79.714 & 77.116 & 76.625 & 74.633 & 66.679 & 62.783 \\ \hline
        ilpd & 0.203 & 0.171 & 0.155 & 0.22 & 0.204 & 0.178 & 0.172 & 0.132 & 0.111 & 0.099 \\ \hline
        mfeat-fourier & 12.745 & 12.761 & 12.749 & 12.89 & 12.66 & 12.456 & 12.165 & 11.485 & 8.031 & 6.47 \\ \hline
        steel-plates-fault & 1.775 & 1.646 & 1.626 & 1.455 & 1.565 & 1.391 & 1.299 & 1.125 & 0.644 & 0.466 \\ \hline
        mfeat-pixel & 93.229 & 94.787 & 95.442 & 89.544 & 84.487 & 72.357 & 64.771 & 55.729 & 35.274 & 16.373 \\
    \bottomrule
    \end{tabular}
\end{small}
\end{center}
\end{table*}

\end{document}